\newif\ifrevmarks
\revmarksfalse

\documentclass[11pt]{article}

\usepackage[final]{acl}

\usepackage{times}
\usepackage{latexsym}
\usepackage{booktabs}
\usepackage[table]{xcolor}
\usepackage{amsmath}
\usepackage{amssymb}
\usepackage{makecell}
\usepackage[most]{tcolorbox}
\newtcolorbox{promptbox}[2][]{%
  enhanced,
  width=\columnwidth,
  colframe=black,
  colback=gray!12,
  colbacktitle=gray!70,
  coltitle=white,
  fonttitle=\bfseries\fontsize{9}{11}\selectfont,
  fontupper=\fontsize{9}{11}\selectfont,
  title=#2,
  arc=2.5mm,
  outer arc=2.5mm,
  boxrule=0.8pt,
  left=2mm,
  right=2mm,
  top=1.2mm,
  bottom=1.2mm,
  boxsep=1mm,
  before skip=6pt,
  after skip=6pt,
  #1
}

\usepackage[T1]{fontenc}

\usepackage[utf8]{inputenc}

\usepackage{microtype}

\usepackage{inconsolata}

\usepackage{graphicx}
\usepackage{pgfplots}
\usepgfplotslibrary{groupplots}
\usetikzlibrary{arrows.meta,decorations.pathreplacing,shapes.geometric,calc}
\usepackage{pifont}
\usepackage{fancyvrb}
\usepackage{soul}
\usepackage{siunitx}

\definecolor{g3poTint}{HTML}{F3EAFC}
\newcommand{\best}[1]{\textbf{#1}}
\newcommand{\second}[1]{\underline{#1}}
\newcommand{\na}{\multicolumn{1}{c}{\textcolor{black!40}{n/a}}}
\newcommand{\unp}{\multicolumn{1}{c}{\textcolor{black!40}{--}}}
\newcommand{\grouprow}[2]{\multicolumn{#1}{l}{\textit{#2}}}
\newcommand{\cmark}{\ding{51}}
\newcommand{\xmark}{\textcolor{black!35}{\ding{55}}}
\ifrevmarks\colorlet{rev}{blue}\else\colorlet{rev}{black}\fi
\newcommand{\add}[1]{\textcolor{rev}{#1}}
\definecolor{chgGreen}{HTML}{00873C}
\newcommand{\chg}[1]{\textcolor{rev}{#1}}
\definecolor{orgOrange}{HTML}{E06A00}
\newcommand{\org}[1]{\textcolor{rev}{#1}}
\newcommand{\fix}[1]{\textcolor{rev}{#1}}
\newcommand{\orng}[1]{\textcolor{rev}{#1}}

\title{G3PO: Leveraging a Hybrid Use of Verifiability and Preference in Reinforcement Learning}

\author{
 \textbf{Massimiliano Pronesti\textsuperscript{1,2}}
 \textbf{Angelo Miculescu\textsuperscript{2}}
 \textbf{Mohsin Kapdi\textsuperscript{2}}
 \textbf{Anya Belz\textsuperscript{2}}
 \textbf{Yufang Hou\textsuperscript{1,3}}
\\
 \textsuperscript{1}IBM Research
 \textsuperscript{2}Dublin City University\\
 \textsuperscript{3}IT:U Interdisciplinary Transformation University Austria
\\
 \small{
   \textbf{Correspondence:} \href{mailto:massimiliano.pronesti@ibm.com}{massimiliano.pronesti@ibm.com}, \href{mailto:yufang.hou@it-u.at}{yufang.hou@it-u.at}
 }
}

\begin{document}
\maketitle
\begin{abstract}
Reinforcement Learning with Verifiable Rewards (RLVR) has shown strong performance in eliciting reasoning in LLMs on tasks with deterministic outcomes, while Reinforcement Learning from Human Feedback (RLHF) has remained a viable option for more subjective, reasoning-based tasks. This is especially true when paired with Process Reward Models (PRMs), which are generally trained on human preference data and are increasingly used to shape policy optimisation. However, the two signals are rarely combined beyond a weighted sum, and little work has investigated whether doing so would lead to more stable training signals. To address this gap, we introduce Group verifiability-Prioritised Preference Policy Optimisation (G3PO), a lexicographic policy optimisation strategy that ranks completions first by verifiable correctness and then by PRM preference. We evaluate G3PO on three tasks: Medical Risk of Bias Assessment, Medical Reasoning, and Mathematics. Across these domains, our hybrid strategy is shown to consistently improve on standard baselines as well as domain-specific alternatives, demonstrating that prioritising correctness while preserving preference order yields stronger and more effective training signals. 
\end{abstract}

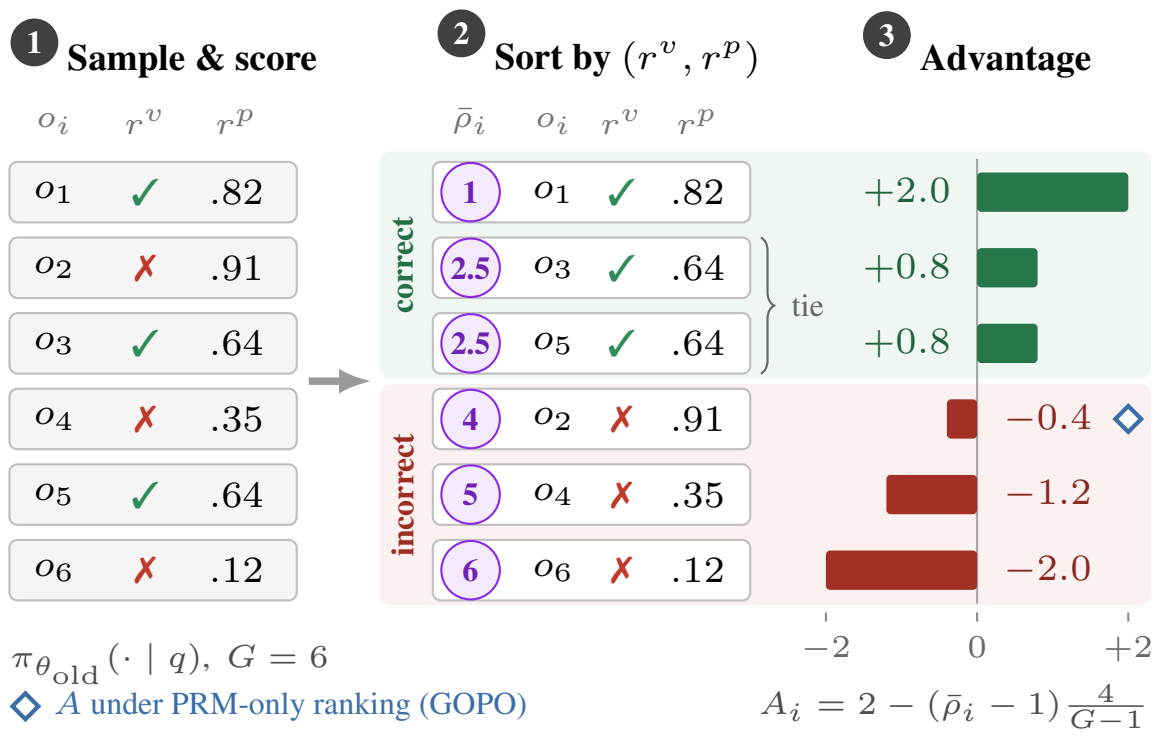
\begin{figure}[t]
    \centering
\begin{tikzpicture}[
    font=\scriptsize,
    x=1cm, y=1cm,
    cell/.style={draw=black!25, fill=black!4, rounded corners=1.5pt,
                 minimum height=0.40cm, inner sep=0pt},
    hdr/.style={font=\scriptsize\bfseries, anchor=south},
    sub/.style={font=\tiny, text=black!55, anchor=south},
    stepnum/.style={circle, fill=black!75, text=white, font=\tiny\bfseries,
                 inner sep=0.6pt, minimum size=9pt},
    rank/.style={circle, draw=g3po, fill=g3po!10, text=g3po!80!black,
                 font=\tiny\bfseries, inner sep=0pt, minimum size=11pt},
]
\definecolor{ok}{HTML}{2E8B57}
\definecolor{bad}{HTML}{C0392B}
\definecolor{g3po}{HTML}{8A2BE2}
\definecolor{gopo}{HTML}{3B6EA8}   
\newcommand{\yes}{\textcolor{ok}{\ding{51}}}
\newcommand{\no}{\textcolor{bad}{\ding{55}}}
\def\zx{6.45}   
\def\sc{0.50}   

\fill[ok!7,  rounded corners=2pt] (2.50, 0.27) rectangle (7.60,-1.23);
\fill[bad!7, rounded corners=2pt] (2.50,-1.27) rectangle (7.60,-2.73);
\node[rotate=90, text=ok!80!black,  font=\tiny\bfseries] at (2.65,-0.48) {correct};
\node[rotate=90, text=bad!80!black, font=\tiny\bfseries] at (2.65,-2.00) {incorrect};

\node[hdr, anchor=south west, inner xsep=0pt] at (0.05,0.62) {\tikz[baseline=-0.6ex]\node[stepnum]{1}; Sample \& score};
\node[hdr] at (3.95,0.62) {\tikz[baseline=-0.6ex]\node[stepnum]{2}; Sort by $(r^v, r^p)$};
\node[hdr] at (6.45,0.62) {\tikz[baseline=-0.6ex]\node[stepnum]{3}; Advantage};
\foreach \x/\t in {0.35/{$o_i$}, 0.95/{$r^v$}, 1.55/{$r^p$},
                   3.10/{$\bar\rho_i$}, 3.65/{$o_i$}, 4.10/{$r^v$}, 4.60/{$r^p$}}
    \node[sub] at (\x,0.25) {\t};

\foreach \k/\o/\v/\p in {0/1/1/.82, 1/2/0/.91, 2/3/1/.64,
                         3/4/0/.35, 4/5/1/.64, 5/6/0/.12} {
    \node[cell, minimum width=1.9cm] at (1.00,-\k*0.5) {};
    \node at (0.35,-\k*0.5) {$o_{\o}$};
    \node at (0.95,-\k*0.5) {\ifnum\v=1 \yes\else\no\fi};
    \node at (1.55,-\k*0.5) {$\p$};
}

\draw[-{Latex[length=2.2mm]}, line width=1.4pt, black!40] (2.03,-1.25) -- (2.45,-1.25);

\draw[black!35] (\zx, 0.22) -- (\zx,-2.72);
\foreach \k/\o/\v/\p/\r/\a in {0/1/1/.82/1/2.0,   1/3/1/.64/2.5/0.8, 2/5/1/.64/2.5/0.8,
                               3/2/0/.91/4/-0.4,  4/4/0/.35/5/-1.2,  5/6/0/.12/6/-2.0} {
    \pgfmathsetmacro{\y}{-\k*0.5}
    \node[cell, minimum width=2.1cm, fill=white] at (3.90,\y) {};
    \node[rank] at (3.10,\y) {\r};
    \node at (3.65,\y) {$o_{\o}$};
    \node at (4.10,\y) {\ifnum\v=1 \yes\else\no\fi};
    \node at (4.60,\y) {$\p$};
    \ifnum\v=1
        \fill[ok!85!black,  rounded corners=0.8pt] (\zx,\y-0.13) rectangle ({\zx+\a*\sc},\y+0.13);
        \node[anchor=east, text=ok!70!black]  at (\zx-0.04,\y) {$+\a$};
    \else
        \fill[bad!85!black, rounded corners=0.8pt] (\zx,\y-0.13) rectangle ({\zx+\a*\sc},\y+0.13);
        \node[anchor=west, text=bad!70!black] at (\zx+0.04,\y) {$\a$};
    \fi
}

\draw[decorate, decoration={brace, amplitude=2.5pt}, black!60]
    (5.02,-0.30) -- (5.02,-1.20) node[midway, right=2pt, font=\tiny, text=black!60] {tie};

\tikzset{gopomark/.style={diamond, draw=gopo, line width=0.8pt, fill=white, inner sep=1.3pt}}
\node[gopomark] at ({\zx+2*\sc},-1.5) {};
\node[gopomark, anchor=west] (lg) at (0.05,-3.40) {};
\node[anchor=west, font=\tiny, text=gopo, inner xsep=2pt] at (lg.east) {$A$ under PRM-only ranking (GOPO)};

\foreach \t/\l in {-2/{$-2$}, 0/{$0$}, 2/{$+2$}}
    \draw[black!45] ({\zx+\t*\sc},-2.78) -- ({\zx+\t*\sc},-2.84)
        node[below=-1pt, font=\tiny, text=black!60, inner xsep=0pt] {\l};

\node[font=\tiny, text=black!70, anchor=north west, inner xsep=0pt] at (0.05,-2.85)
    {$\pi_{\theta_{\mathrm{old}}}(\cdot\mid q)$, \ $G=6$};
\node[font=\tiny, text=black!70, anchor=north east, inner xsep=0pt] at (7.60,-3.12)
    {$A_i = 2-(\bar\rho_i-1)\tfrac{4}{G-1}$};
\end{tikzpicture}
    \caption{G3PO on a group of $G{=}6$ completions. \textbf{(1)}~Each completion receives a verifiable reward $r^v$ (\ding{51}/\ding{55}) and a PRM score $r^p$. \textbf{(2)}~Completions are sorted lexicographically, by correctness first and PRM score second; completions with identical $(r^v, r^p)$ share their mean rank $\bar\rho$ ($o_3$, $o_5$). \textbf{(3)}~Ranks are mapped linearly to advantages in $[-2, 2]$. Every correct completion outranks every incorrect one: $o_2$ has the highest PRM score but is wrong, so G3PO assigns it $A{=}{-0.4}$, whereas ranking by PRM score alone, as in GOPO (\textcolor[HTML]{3B6EA8}{$\diamond$}), would give it the maximum advantage of $+2$.}
    \label{diagram_of_algorithm_workflow}
\end{figure}

\section{Introduction}

Reinforcement Learning (RL) has seen many recent developments that have propelled it to the forefront of modern AI research and LLM post-training. A major development in this direction was Reinforcement Learning from Human Feedback (RLHF), which uses learned preference signals to guide policy optimisation and has been effective in producing more stable outputs that better match human judgment~\cite{Ouyang_RLHF}. Building on this idea, Process Reward Models (PRMs) extend preference-based supervision from the final answer to the intermediate steps~\cite{Lightman_math500}. This gives greater control over the training process, and is especially useful in settings where multiple reasoning paths may lead to the same final answer. Still, some paths are more efficient, coherent, or reliable than others. Moreover, it remains an expensive endeavour to obtain high-quality human annotations, and the lack of an objective signal still leaves the reward models vulnerable to reward hacking~\cite{zhang_lessons_2025}.

Reinforcement Learning with Verifiable Rewards (RLVR) was proposed as a robust alternative that greatly mitigated this issue. By replacing the preference reward with a strict, directly verifiable one, this technique yielded large performance increases in tasks with deterministic outcomes, such as maths or coding~\cite{yue2026does, guo_deepseek-r1_2025, lambert2024tulu}. The natural next step was then to migrate from the outcome and back to the process. Introducing deterministic checks at the step-level helped to further enhance performance in many LLM tasks~\cite{wang2025masked} and eventually led to the introduction of Verifiable Process Reward Models (VPRMs)~\cite{pronesti2026beyond}, which took advantage of complex tasks where processes included easily checkable steps. While proving to be a highly competent method for RL training, this approach has an important limitation: many tasks do not provide verifiable reasoning traces, or traces that are prohibitively expensive to obtain. This indicates that verifiability alone is not always sufficient, and preference-based supervision remains important.

Recent work~\cite{choi_gopo_2026} demonstrated another important use of this preference-based supervision by constructing a rank-based policy optimisation method that ignores reward magnitudes and instead computes policy advantages from the relative ordering of the same rewards. This work culminated in the technique known as Group Ordinal Policy Optimisation (GOPO), an ordinal variant of Group Relative Policy Optimisation (GRPO) \cite{shao_deepseekmath_2024} which traditionally calculated a group-relative advantage using a $z$-score formulation. GOPO outperformed GRPO on tasks that depend more on subjective judgement, suggesting that rank-based optimisation is particularly well suited to preference-trained reward models. At the same time, even complex verifiable tasks require substantial reasoning in order to arrive at the correct answer, and high-quality performance in such settings depends not only on the final outcome but also on the reasoning process itself. This motivates a training objective that prioritises verifiable correctness while still exploiting preference-based ranking among candidates.

To this end, we propose Group Verifiability-Prioritised Preference Policy Optimisation (G3PO), a novel lexicographical advantage estimation strategy that leverages both verifiability and preference-based ordering. G3PO first partitions sampled completions according to correctness, 
then 
applies
a reward model ranking completions within those partitions, in a manner analogous to GOPO. Correct completions are always prioritised, while the preference information is used to distinguish among candidates with the same correctness status. In this way, G3PO preserves the correctness structure of verifiable tasks while still encouraging higher-quality reasoning trajectories. Figure~\ref{diagram_of_algorithm_workflow} shows a visualisation of the workflow.

To evaluate this framework, we apply our method to three challenging, real-world tasks, all of which contain verifiability but also heavy reasoning: Risk of Bias (RoB) assessment in clinical systematic reviews, Mathematics, and General Medical Reasoning. We compare against GRPO, GOPO, and strong domain-specific baselines such as MedReason~\cite{medreason} and NFT~\cite{NFT}\add{, as well as weighted-sum and decoupled-normalisation (GDPO;~\citealp{liu2026gdpo}) combinations of the same two rewards}. Across these settings, we demonstrate that G3PO is consistently competitive with, and in most cases outperforms, the comparison methods.

In summary, our contributions are as follows:
\begin{enumerate}
    \item We introduce G3PO, a novel policy optimisation framework for complex verifiable tasks that combines correctness-first ordering with preference-based ranking.
    \item We evaluate G3PO on three challenging task domains, each comprised of a series of internal benchmarks, and compare it against GRPO, GOPO, and strong domain-specific baselines.
    \item \add{We show that the gains come from the lexicographic order itself: G3PO outperforms tuned weighted-sum and GDPO combinations of the same rewards, limits the cost of PRM exploitation under extended training, and transfers to rubric-graded open-ended medical questions.}
\end{enumerate}

\section{Related Work}

\subsection{From Human Feedback to Verifiable Rewards}

Post-training has shifted from reinforcement learning from human feedback (RLHF), where a learned reward model trained on human preferences guides Proximal Policy Optimisation (PPO) updates \cite{Ouyang_RLHF,schulman_proximal_2017}, toward reinforcement learning with verifiable rewards (RLVR), where rewards come directly from ground-truth checks. This shift was driven by reward hacking in reward models \cite{amodei_concrete_2016,hasselt_deep_2018} and by domains like mathematics, where correctness is unambiguous.

DeepSeekMath \cite{shao_deepseekmath_2024} introduced Group Relative Policy Optimisation (GRPO), replacing the critic network with relative reward ranking across sampled completions, which reduced the computational overhead compared to PPO. DeepSeek-R1~\cite{guo_deepseek-r1_2025} showed this method could even induce strong emergent reasoning purely through RL. However, normalising by group statistics makes its advantage depend on the group's composition: groups with identical rewards give no signal, and a lone correct completion receives an advantage that grows with $G$.

Later work refined GRPO: Decoupled Clip and Dynamic sAmpling Policy Optimisation (DAPO)~\cite{yu2025dapo} fixed entropy collapse via decoupled clipping and dynamic sampling, while Group Ordinal Policy Optimisation (GOPO)~\cite{choi_gopo_2026} replaced scalar advantage with ranking-based optimisation for noisy rewards. \citet{OaT-Zero} identified a length bias in GRPO that inflates response length for incorrect outputs and introduced Dr.\ GRPO, an unbiased variant that improves token efficiency without sacrificing accuracy. \citet{liu2026gdpo} further showed that applying GRPO under multi-reward settings collapses distinct reward combinations into identical advantage values, and proposed GDPO, which decouples the normalisation of individual rewards to preserve training signal resolution. Whether RLVR truly expands reasoning capacity remains debated:  \citet{yue2026does} find it mainly boosts sampling efficiency rather than creating new reasoning paths, whereas  \citet{wen2025reinforcement} argue this is a metric artefact and show genuine gains under a reasoning-aware metric, with~\citet{wang2025masked} offering a related account of credit propagation. A parallel question is generalisation beyond maths/code: \citet{su2025crossing} extend RLVR to medicine via a soft, model-based reward signal, and \citet{NFT} similarly blend supervised and RL signals toward this end, directly relevant to applying GRPO-style objectives to medical reasoning.

\subsection{Process Reward Models versus Outcome Verification}

A complementary thread targets reward granularity: since outcome models can reward correct answers reached through flawed logic, process reward models (PRMs) supervise individual steps for more reliable signals \cite{Lightman_math500}. Math-Shepherd \cite{wang_math-shepherd_2024} automated this without human labels; later work exposed step-level reward hacking and pushed toward more structured step representations \cite{pronesti2026beyond, zhang_lessons_2025}. Med-PRM \cite{medprm} extended this to medicine, grounding step rewards in clinical guidelines.

\section{Background}



\subsection{Group Relative Policy Optimisation}

Often seen as the successor to Proximal Policy Optimisation (PPO) \cite{schulman_proximal_2017} as the most widely adopted policy optimisation strategy for Reinforcement Learning, GRPO \cite{shao_deepseekmath_2024} discards the resource-intensive critic model used in PPO and instead compares multiple candidate completions for the same input.

More precisely, for each question $q$, the old policy $\pi_{\theta_{old}}$ samples a group of G candidate completions $\{o_1, o_2,...,o_G\}$. These completions are then used to update the main policy model $\pi_\theta$ by maximising the following objective function:

\begin{multline*}\label{grpo_main}
\mathcal{J}_{\mathrm{GRPO}}(\theta)
=
\mathbb{E}\Bigg[
\frac{1}{G}
\sum_{i=1}^{G}
\frac{1}{|o_i|}
\sum_{l=1}^{|o_i|}
\min\Big(
p_{i,l}(\theta)A_i, \\
\mathrm{clip}\big(p_{i,l}(\theta),1-\epsilon,1+\epsilon\big)A_i
\Big)
-
\beta\,\mathbb{D_{\mathrm{KL}}}\big[\pi_\theta \,\|\, \pi_{\mathrm{ref}}\big]
\Bigg]
\end{multline*}

where $\epsilon$ is the hyperparameter that governs the level of clipping, $p_{i,l}(\theta)$ is the token-level probability ratio defined as:

\begin{equation*}
p_{i,l}(\theta)
=
\frac{\pi_\theta(o_{i,l} \mid q, o_{i,<l})}{\pi_{\mathrm{\theta_{old}}}(o_{i,l} \mid q, o_{i,<l})},
\end{equation*}

and the $\mathrm{KL}$ regularisation term, whose strength is governed by $\beta$, is computed using Schulman's approximation~\cite{schulman}:

\begin{equation*}
D_{\mathrm{KL}}(\pi_\theta \| \pi_{\mathrm{ref}})
\approx
\frac{\pi_{\mathrm{ref}}(o_i \mid q)}{\pi_\theta(o_i \mid q)}
-
\log \frac{\pi_{\mathrm{ref}}(o_i \mid q)}{\pi_\theta(o_i \mid q)}
- 1
\end{equation*}

Here, $\pi_{ref}$ is the frozen reference policy, usually pre-trained using Supervised Fine-Tuning (SFT).

Lastly, the advantage terms $A_i$ are computed by comparing the grouped rewards $\{r_1, r_2,...,r_G\}$ corresponding to the outputs for each set of questions. The calculation is performed using the standard $z$-score formula:

\begin{equation*}
    A_i^\mathrm{GRPO} = \frac{r_i - \mathrm{mean}\left(\{r_1, r_2, \cdots, r_G\}\right)}{\mathrm{std}\left(\{r_1, r_2, \cdots, r_G\}\right)}.
\end{equation*}

\subsection{Group Ordinal Policy Optimisation}

Group Ordinal Policy Optimisation (GOPO)~\cite{choi_gopo_2026} makes use of the same overall objective function $\mathcal{J}_{\mathrm{GRPO}}(\theta)$, but it differs in its advantage estimation formulation. GOPO uses only the ranking of the rewards and discards their magnitudes. This method is designed to work specifically with PRMs, which are trained on preference data and so are better equipped to distinguish between superior completions rather than applying a specific reward to each. The most important factor is the specific order (rank) of the completions. It computes, therefore, the advantage as follows:

\begin{equation*}
    A_i^{\mathrm{GOPO}} = 2 - \{\rho(i) - 1\}\frac{4}{G - 1}
\end{equation*}

where $\rho : \{o_1,o_2,...,o_G\}\rightarrow [G]$ is the rank mapping of each completion $o_i$, according to its reward $r_i$, i.e.\ the completion with the highest rank receives rank 1 and the lowest receives rank $G$. The upper and lower bounds of $A_i^{GOPO}$ align with 2 standard deviations of the standardised variables, and the advantages are spaced $\tfrac{4}{G-1}$ apart.



\section{Methodology}

In this work, we consider the scenario where each completion $o_i$ is evaluated by a tuple of rewards $(r^v_i , r^p_i )$, representing a verifiable reward and a process reward model (PRM) score, respectively. To bridge the gap between magnitude-sensitive verifiable outcomes and ordinal process signals, we propose the following distinct advantage estimation strategy.

\subsection{Group verifiability-Prioritised Preference Policy Optimisation (G3PO)}

The method extends the GOPO framework to a multi-objective ordinal
setting. We establish a lexicographic preference relation $\succ$ that strictly prioritises verifiable success:

{\small
\begin{equation*}
\begin{aligned}
\left(r_i^v, r_i^p\right) &\succ \left(r_j^v, r_j^p\right) \iff  \left(r_i^v > r_j^v\right) \lor \left(r_i^v = r_j^v \land r_i^p > r_j^p\right)
\end{aligned}
\end{equation*}
}

To ensure permutation invariance and prevent the injection of spurious gradient noise, we employ a symmetric rank \(\bar{\rho}(i)\). Let
$E_i = \left\{ j \in [G] \mid r_j^v = r_i^v \wedge r_j^p = r_i^p \right\}$ be the equivalence class of indices with identical reward profiles. The symmetrised rank is defined as the mean of the strict ranks \orng{under $\succ$, ties broken in any order,} spanned by the group:
\begin{equation*}
\bar{\rho}(i) = \frac{1}{|E_i|} \sum_{j \in E_i} \rho(j)
\end{equation*}

The hybrid advantage is then computed following the linear mapping:

\begin{equation*}\label{eq8}
A_i^{\mathrm{G3PO}} = 2 - \{\bar{\rho}(i) - 1\}\frac{4}{G - 1}
\end{equation*}

The formulation of $E_i$ and the subsequent symmetric rank $\bar{\rho}(i)$ ensure that in the case of a tie between two completions in both verifiability and preference, both would receive the same advantage.

\subsection{\texorpdfstring{\add{Properties of the Advantage}}{Properties of the Advantage}}\label{sec:properties}

\subsubsection{Relation to GOPO}

Following~\citet{choi_gopo_2026}, \fix{consider their smoothed single-prompt objective and} let \(\xi\) denote the centred gradient \fix{given the advantages. If the centred score vectors are conditionally uncorrelated with second moment $\sigma_X^2$, then}
{\color{rev}\setlength{\abovedisplayskip}{4pt}\setlength{\belowdisplayskip}{4pt}
\begin{equation*}
\mathbb{E}\big[\|\xi\|^2 \mid A\big]
=
\frac{\sigma_X^2}{G}\cdot\frac{1}{G}\sum\nolimits_{i=1}^{G}A_i^2,
\end{equation*}%
}%
so the \fix{gradient variance} is proportional to the second moment of the transformed advantages. \fix{For rank advantages, it exceeds GRPO's at every $G$, most of all at small $G$ (Section~\ref{sec:order}).} \fix{However}, the bounded rank transformation ensures that GOPO's \chg{advantages stay within a range that does not depend on} \(G\), in contrast to GRPO, \chg{whose largest} \fix{attainable} \chg{advantage grows with $G$}.

G3PO inherits the same boundedness property. Since its \fix{symmetrised} lexicographic rank satisfies \(\bar{\rho}(i)\in[1,G]\), its advantage lies in \(A_i^{\mathrm{G3PO}}\in[-2,2]\). As a result, the corresponding centred gradient remains uniformly controlled with respect to \(G\)\fix{: $\mathbb{E}[\|\xi\|^2\mid A]\leq 4\sigma_X^2/G$}.

%

\subsubsection{\texorpdfstring{\chg{Ordering and Bound}}{Ordering and Bound}}\label{sec:order}

\chg{Because $r^v$ is compared first, G3PO} \fix{gives every correct completion a larger advantage than every incorrect one.} \chg{A weighted sum} \fix{$\alpha r^v+(1-\alpha)r^p$} \chg{guarantees this} \fix{in every group} \chg{only if} \fix{$\alpha$ exceeds $(1-\alpha)$ times the range of $r^p$ ($\alpha>0.5$ for $r^p\in[0,1]$)}\chg{, and normalising each reward separately, as GDPO does, gives no such guarantee. G3PO's advantages lie in $[-2,2]$ for every $G$, whereas GRPO's are bounded by $(G-1)/\sqrt{G}$ (sample standard deviation), which exceeds $2$ for $G\geq6$. The bound does not make the} \fix{update less noisy}\chg{: without ties, $\frac{1}{G}\sum_i (A_i^{\mathrm{G3PO}})^2=\frac{4(G+1)}{3(G-1)}$, $1.71$ at $G=8$, compared to $(G-1)/G$ for GRPO, and only ties lower it. What it removes is the large advantage GRPO assigns to a lone correct completion in a group of incorrect ones.}

\section{Experimental Setup}


All training was conducted through RL with no prior SFT. The overall objective function used for policy optimisation was GRPO~\cite{shao_deepseekmath_2024} with three algorithmic variations: GRPO's studentisation score, GOPO's ranking formulation~\cite{choi_gopo_2026}, and our own G3PO hybrid advantage.

\subsection{Policy Models}

Training was conducted with the following two 7B parameter models: Qwen2.5-7B-Instruct~\cite{qwen2.5}, used for Risk of Bias Assessment \& Medical Reasoning, and Qwen2.5-7B-Math~\cite{qwen2.5Math}, as is consistent with other methods evaluating on Math tasks. Models were trained for 1 epoch, with 8 generated completions per question and learning rates ranging from $5.0 \times 10^{-7}$ for medical tasks and Maths, and $3.0 \times 10^{-6}$ for RoB Assessment. The maximum generated steps were capped at $2048$ for all domains. Additional details are reported in Appendix~\ref{app:hyperparams}.

\begin{table*}[t]
\centering
\setlength{\tabcolsep}{2.7pt}
\renewcommand{\arraystretch}{1.08}
\small
\begin{tabular}{l *{3}{S[table-format=2.1]} S[table-format=2.1] @{\hspace{12pt}} *{4}{S[table-format=2.1]} S[table-format=2.1]}
\toprule
& \multicolumn{4}{c}{\textbf{Risk of Bias}} & \multicolumn{5}{c}{\textbf{Medical Reasoning}} \\
\cmidrule(lr){2-5} \cmidrule(l){6-10}
\textbf{Model}
& {\makecell{Cochrane-\\Forest}} & {\makecell{RoBBR\\Cochrane}} & {\makecell{RoBBR\\Non-Coch.}} & {\textbf{Avg}}
& {MedQA} & {\makecell{Med-\\MCQA}} & {\makecell{MMLU-Pro\\Health}} & {\makecell{MMLU-Pro\\Biology}} & {\textbf{Avg}} \\
\midrule
\grouprow{10}{Pretrained Models} \\
\addlinespace[1pt]
Qwen2.5-7B-Instruct    & 32.9 & 35.8 & 36.4 & 35.0 & 59.8 & 54.2 & 56.5 & 69.3 & 60.0 \\
Qwen2.5-14B-Instruct   & 39.0 & 37.0 & 35.4 & 37.1 & 59.6 & {\best{60.9}} & {\best{61.3}} & 65.6 & 61.8 \\
DeepSeek-Qwen-7B       & \unp & \unp & \unp & \unp & 18.1 & 19.7 & 11.9 & 30.8 & 20.1 \\
DeepSeek-Qwen-14B      & 33.3 & 35.8 & 35.4 & 34.8 & 38.4 & 34.9 & 48.0 & 46.9 & 42.1 \\
AlphaMed-7B-base-RL    & 46.1 & 37.2 & 36.8 & 40.0 & 60.8 & 58.5 & 58.1 & 68.4 & 61.5 \\
HuatuoGPT-o1-8B        & 30.7 & 31.6 & 29.8 & 30.7 & 24.2 & 37.0 & 30.0 & {\second{73.9}} & 41.3 \\
Qwen2.5-7B-VPRM-GRPO\footnotemark & {\best{76.7}} & {\best{58.5}} & {\best{47.2}} & {\best{60.8}} & \na & \na & \na & \na & \na \\
Qwen2.5-7B-VPRM-DAPO   & 60.6 & 48.9 & 45.3 & 51.6 & \na & \na & \na & \na & \na \\
\midrule
\grouprow{10}{Qwen2.5-7B trained with RL (this work)} \\
\addlinespace[1pt]
\quad GRPO & {\add{70.4}} & {\add{\second{57.8}}} & {\add{45.0}} & {\add{57.7}} & {\add{\second{64.4}}} & {\add{57.1}} & {\add{57.6}} & {\add{71.7}} & {\add{\second{62.7}}} \\
\quad GOPO & {\add{68.3}} & {\add{52.8}} & {\add{42.9}} & {\add{54.7}} & \unp & \unp & \unp & \unp & \unp \\
\rowcolor{g3poTint}
\quad \textbf{G3PO (ours)} & {\add{\second{74.8}}} & {\add{57.7}} & {\add{\second{45.9}}} & {\add{\second{59.5}}} & {\add{\best{65.8}}} & {\add{\best{60.9}}} & {\add{\second{59.9}}} & {\add{\best{74.1}}} & {\add{\best{65.2}}} \\
\bottomrule
\end{tabular}
\caption{Accuracy (\%) on risk-of-bias assessment and medical reasoning benchmarks. Averages are computed separately per domain. Best scores are in \textbf{bold}, second best \underline{underlined}. ``n/a'': not applicable, since VPRM training requires step-level verifiable supervision, which is unavailable for medical reasoning; ``--'': unparsable or inconclusive outputs.}
\label{tab:medical_benchmarks}
\end{table*}

\subsection{Task Rewards}

The key rewards used in all tasks include formatting, correctness, and a Process Reward Model (PRM). For the main experiments, G3PO was the only formulation that used all three rewards. GRPO was trained using formatting and correctness signals to isolate its verifiability component, which also adheres to its standard usage. In contrast, GOPO was provided with formatting and PRM rewards to capture the preference-based component, consistent with its internal design. Section~\ref{section/ablations} details reward ablations completed for GRPO and GOPO, used to compare against our own G3PO. The PRMs used included Qwen2.5-Math-PRM-7B~\cite{zhang_lessons_2025} for Maths tasks and Llama-3.1-medprm-reward-v1.0 \cite{medprm} for both RoB and Med tasks. The latter was chosen for its strong internal clinical expertise and built-in retrieval-augmented generation (RAG) capabilities \cite{medprm}.

{\color{rev}To separate the lexicographic order from the use of two rewards, we also train with both signals combined as a weighted sum \(R_i=\alpha r^v_i+(1-\alpha)r^p_i\), \(\alpha\in\{0.2,0.5,0.8\}\), and with GDPO~\cite{liu2026gdpo}, which normalises each reward separately before summing\orng{, preserving each reward's relative differences}. As GRPO's advantage is scale-invariant, \(\alpha=0.5\) equals the unweighted sum.}

\subsection{Tasks and Datasets}

\subsubsection{Risk of Bias}
Risk of bias assessment is the critical task in medical systematic reviews of evaluating clinical trials for methodological flaws that could systematically distort their reported results. For this task, we consider three datasets. First,~\textsc{CochraneForest}~\cite{pronesti_enhancing_2025, pronesti_query-driven_2025}, containing $1,846$ annotated examples drawn from 48 Systematic Reviews conducted by the Cochrane network of researchers.
The second consists of the two test sets from the RoBBR benchmark: RoBBR Cochrane, which contains 906 datapoints originating from 204 papers included in 58 Cochrane reviews; and RoBBR–NonCochrane, which contains 2,489 datapoints drawn from 496 non-Cochrane reviews that collectively assess 496 papers.

In both subsets, item structures contain the bias to be considered, the PICO (Population, Intervention, Comparison, Outcome), objective, the full paper, and the final label, of which the only available options are low (\textbf{--}), high (\textbf{+}), and unclear (\textbf{?}). 

\subsubsection{Medical Reasoning}

The motivation for selecting this domain was to evaluate our framework in a manner analogous to RoB assessment, but with a broader focus on general medical reasoning rather than specifically considering bias as the primary target variable. Our approach differed from standard domain procedures in that it lacked any SFT, while focusing solely on RL training. Our models were trained using m23k-tokenised~\cite{m23k}, a collection of 23.5k multiple-choice medical questions taken from the training sets of other popular datasets.

In accordance with relevant evaluation frameworks, and due to each of them containing a distinct output that is easily verifiable, we tested our models on the following datasets: MedQA~\cite{medqa}, MedMCQA~\cite{medmcqa}, MMLU-Pro~\cite{wang2024mmlu}. The latter is divided into subsets of varying topics (i.e.\ Law, Business, Maths, Science). For this experiment, focusing on general medicine, we selected just the Health \& Biology splits. 

\subsubsection{Maths}
Mathematics is one of the most heavily benchmarked domains in LLM reasoning, valued for its clear correctness criteria and unambiguous evaluation of model performance. To study this setting, we trained on MATH~\cite{hendrycks2021measuring_mathdataset}, a large-scale mathematics dataset containing diverse problem-solving traces, and evaluated on a suite of benchmarks chosen to cover a range of difficulty levels and reasoning styles. This includes GSM8K~\cite{cobbe_gsm8k} and MATH500~\cite{Lightman_math500}, which test grade school and competition-style mathematical reasoning, as well as AIME24~\cite{aime24}, AIME25~\cite{aime25}, AIME26~\cite{aime26}, Olympiad~\cite{olympiadbench} and AMC23~\cite{amc23}, which provide progressively more challenging problems that better probe advanced reasoning and generalisation.

\section{Results}\label{sec:results}

The following section details the evaluation conducted on our own trained models as well as on relevant baselines and domain-specific alternative LLMs. We report \add{accuracy averaged over $k$ sampled answers per question (avg@$k$),} using $k=4$ for RoB tasks and $k=32$ for both Medical Reasoning and Maths.

\footnotetext{The slight performance advantage of the VPRM-trained models is expected, given its narrow optimisation for RoB and step-level verifiable supervision.
}

\subsection{Task Domain 1: Risk of Bias}
Table \ref{tab:medical_benchmarks} (left) reports results obtained on the Risk of Bias benchmarks, split between models trained by ourselves versus those that were pre-trained and run solely in zero-shot. Among these,~\citet{pronesti2026beyond}'s VPRM-trained using GRPO achieves the strongest performance across all test sets. This is consistent with how their framework makes strong use of verifiable reward signals coming from each step in the process, rather than solely from the outcome. Nevertheless, G3PO remains consistently competitive, falling short by no more than $2\%$ in accuracy across the benchmarks, despite being a more general method rather than one tailored specifically to one use case.

These results also suggest that RoB-specific pre-training provides a clear advantage over more general medical reasoning models. The general baselines, including AlphaMed-7B-base-RL~\cite{alphamed} and HuatuoGPT-o1-8B~\cite{chen2024huatuogpt}, perform substantially worse, as do the Qwen2.5 and DeepSeek-Qwen baselines. Furthermore, DeepSeek-Qwen-7B produces no parsable results on these benchmarks, underscoring the difficulty of transferring general reasoning models to RoB assessment without task-specific adaptation.

Across our other methods, GRPO and GOPO both substantially outperform the pre-trained non-specialised baselines, with GRPO even achieving a slightly higher score than G3PO on one of the RoB benchmarks\add{ (RoBBR Cochrane: 57.8 vs.\ 57.7, within seed variance)}. G3PO, however, is the most consistent across the full set of benchmarks, matching the strongest pre-trained model closely while remaining competitive throughout.

\begin{table*}[t]
\centering
\setlength{\tabcolsep}{6pt}
\renewcommand{\arraystretch}{1.08}
\small
\begin{tabular}{l S[table-format=2.1] S[table-format=2.1] S[table-format=2.1] *{2}{S[table-format=1.1]} S[table-format=2.1] S[table-format=2.1] @{\hspace{14pt}} S[table-format=2.1]}
\toprule
& & & \multicolumn{3}{c}{\textbf{AIME}} & & & \\
\cmidrule(lr){4-6}
\textbf{Model} & {GSM8K} & {MATH500} & {'24} & {'25} & {'26} & {Olympiad} & {AMC23} & {\textbf{Avg}} \\
\midrule
\grouprow{9}{Pretrained Models} \\
\addlinespace[1pt]
Qwen2.5-Math-7B (base)    & 67.2 & 53.4 &  9.8 & 2.3 & 1.9 & 20.9 & 40.8 & 28.0 \\
DeepSeek-Math-7B-RL       & 69.8 & 45.7 & 16.7 & 3.8 & 1.9 & 18.1 & 23.6 & 25.7 \\
Eurus-2-7B-PRIME-Zero     & 63.4 & 58.6 &  9.0 & 4.1 & 2.4 & 23.4 & 43.6 & 29.2 \\
Qwen2.5-Math-7B-OaT-Zero  & 90.2 & 68.9 & {\second{20.7}} & 5.2 & {\second{5.1}} & 35.1 & 56.5 & 40.2 \\
NFT-7B                    & {\best{91.8}} & {\second{70.2}} & 18.8 & {\second{6.4}} & 3.7 & {\second{35.7}} & {\best{62.3}} & {\second{41.3}} \\
\midrule
\grouprow{9}{Qwen2.5-Math-7B trained with RL (this work)} \\
\addlinespace[1pt]
\quad GRPO & 83.4 & 64.3 & 18.6 & 5.5 & 4.2 & 32.6 & 52.9 & 37.4 \\
\quad GOPO & 80.7 & 63.9 & 16.1 & 5.4 & 4.9 & 32.8 & 56.2 & 37.1 \\
\rowcolor{g3poTint}
\quad \textbf{G3PO (ours)} & {\second{90.5}} & {\best{70.3}} & {\best{22.7}} & {\best{7.8}} & {\best{6.5}} & {\best{36.4}} & {\second{60.7}} & {\best{42.1}} \\
\bottomrule
\end{tabular}
\caption{Accuracy (\%) on mathematical reasoning benchmarks. Best scores are in \textbf{bold}, second best \underline{underlined}.}
\label{tab:math_benchmarks}
\end{table*}

\subsection{Task Domain 2: Medical Reasoning}

The remainder of Table~\ref{tab:medical_benchmarks} (right) reports results obtained for Medical Reasoning benchmarks. These results highlight an important limitation of methods that rely on step-level reward signals such as VPRMs: when such supervision is unavailable, this training strategy cannot be applied. Outcome verifiability remains a strong training signal, with our GRPO-trained model achieving strong performance on all benchmarks. However, G3PO's hybrid approach continues to report the highest scores across three of the four benchmarks\add{ (tied with Qwen2.5-14B-Instruct on MedMCQA)} and remains competitive in the last.

The results obtained by the pre-trained reasoning baselines show a clear dependence on model scale. In all cases, larger variants of both Qwen2.5 and DeepSeek-Qwen outperform their small counterparts, indicating that strong factual recall, a large component of this task, is more easily attainable in models with more parameters and therefore more capacity. Among domain-specific alternatives, AlphaMed performs best, with HuatuoGPT-o1 remaining competitive only on one benchmark.

\subsection{Task Domain 3: Maths}

Table~\ref{tab:math_benchmarks} reports our results across the various Maths benchmarks. Across the three training algorithms, G3PO consistently outperforms the other two on every evaluated benchmark, attaining the highest overall scores in five of seven benchmarks. An additional observation is that neither GRPO, which is trained with a verifiable reward, nor GOPO, which is trained with a PRM, demonstrates a consistent advantage over the other across the maths benchmarks. On the more standard datasets such as GSM8K~\cite{cobbe_gsm8k} and MATH500~\cite{Lightman_math500}, which contain grade and secondary school level questions, GRPO performs slightly better, suggesting that verifiable reward signals are particularly effective in this setting. By contrast, on the more challenging benchmarks, including AIME24/25/26~\cite{aime24, aime25, aime26}, Olympiad~\cite{olympiadbench}, and AMC23~\cite{amc23}, the two methods are much closer in performance, and in some cases GOPO is marginally stronger. Overall, this indicates that the relative benefit of verifiability versus preference-based training is not uniform across the full suite of benchmarks and may be affected by the overall benchmark difficulty.

We also compare against four powerful pre-trained, maths-specific alternatives: Qwen2.5-Math-7B-OaT-Zero~\cite{OaT-Zero}, Eurus-2-7B-PRIME-Zero~\cite{Eurus-2-7B}, DeepSeek-Math-7B-RL~\cite{shao_deepseekmath_2024}, and NFT-7B~\cite{NFT}. Among these models, NFT-7B achieves the strongest pre-trained performance overall, obtaining the top score on two benchmarks, with OaT-Zero consistently maintaining pace and outperforming it on a couple of metrics. Overall, however, G3PO attains the most consistent results even compared to these models, indicating that the proposed training procedure yields the strongest overall performance in this domain.

\begin{table*}[t]
\centering
\setlength{\tabcolsep}{4.2pt}
\renewcommand{\arraystretch}{1.08}
\small
\begin{tabular}{l ccc S[table-format=2.1] S[table-format=2.1] S[table-format=2.1] *{2}{S[table-format=1.1]} S[table-format=2.1] S[table-format=2.1] @{\hspace{10pt}} S[table-format=2.1]}
\toprule
& \multicolumn{3}{c}{\textbf{Reward}} & & & \multicolumn{3}{c}{\textbf{AIME}} & & & \\
\cmidrule(lr){2-4} \cmidrule(lr){7-9}
\textbf{Advantage} & $r^v$ & $r^p$ & \add{$\alpha$} & {GSM8K} & {MATH500} & {'24} & {'25} & {'26} & {Olympiad} & {AMC23} & {\textbf{Avg}} \\
\midrule
GRPO & \cmark & \xmark & \add{--} & 83.4 & 64.3 & {\second{18.6}} & 5.5 & 4.2 & 32.6 & 52.9 & 37.4 \\
     & \xmark & \cmark & \add{--} & 78.3 & 67.9 & 13.4 & 5.3 & 3.5 & 31.2 & 55.3 & 36.4 \\
     & \cmark & \cmark & \add{0.5} & 88.4 & 67.8 & 14.3 & 4.8 & 4.0 & 29.3 & 53.9 & 37.5 \\
     & \add{\cmark} & \add{\cmark} & \add{0.8} & {\add{87.9}} & {\add{\second{68.1}}} & {\add{18.5}} & {\add{\second{6.2}}} & {\add{\second{5.0}}} & {\add{\second{34.1}}} & {\add{\second{56.5}}} & {\add{\second{39.5}}} \\
     & \add{\cmark} & \add{\cmark} & \add{0.2} & {\add{\second{89.4}}} & {\add{59.8}} & {\add{14.1}} & {\add{4.7}} & {\add{3.2}} & {\add{29.1}} & {\add{51.4}} & {\add{36.0}} \\
\midrule
GOPO & \xmark & \cmark & \add{--} & 80.7 & 63.9 & 16.1 & 5.4 & 4.9 & 32.8 & 56.2 & 37.1 \\
     & \cmark & \xmark & \add{--} & 82.2 & 63.4 & 15.8 & 5.4 & 3.7 & 31.3 & 50.2 & 36.0 \\
\midrule
\add{GDPO} & \add{\cmark} & \add{\cmark} & \add{--} & {\add{87.7}} & {\add{67.9}} & {\add{14.5}} & {\add{4.9}} & {\add{4.4}} & {\add{30.1}} & {\add{54.0}} & {\add{37.6}} \\
\midrule
\rowcolor{g3poTint}
\textbf{G3PO} & \cmark & \cmark & \add{--} & {\best{90.5}} & {\best{70.3}} & {\best{22.7}} & {\best{7.8}} & {\best{6.5}} & {\best{36.4}} & {\best{60.7}} & {\best{42.1}} \\
\bottomrule
\end{tabular}%
\caption{Reward and advantage ablations on mathematical reasoning (Qwen2.5-Math-7B). $r^v$: verifiable outcome reward; $r^p$: PRM score. \add{$\alpha$: weight of $r^v$ in the weighted sum $\alpha r^v+(1-\alpha)r^p$ ($0.5$: unweighted sum); GDPO normalises each reward separately before summing.} The first row of each block is the default configuration used in the main results. Naively combining both signals brings little or no gain \org{under GRPO}, while G3PO's lexicographic combination improves on every benchmark.}
\label{ablations}
\end{table*}

\begin{figure*}[t]
    \centering
\definecolor{colorG3PO}{HTML}{8A2BE2}
\definecolor{colorGRPO}{HTML}{FF8C00}
\definecolor{colorGDPO}{HTML}{1A9E77}
\newcommand{\stabstep}[3]{%
  \addplot[color=#3, opacity=0.22, line width=0.4pt, forget plot]
    table[x=step, y=raw, col sep=comma] {diagrams/grad_stab/#1_steps_#2.csv};
  \addplot[color=#3, forget plot] table[x=step, y=ema, col sep=comma] {diagrams/grad_stab/#1_steps_#2.csv};
}
\newcommand{\stabG}[3]{%
  \addplot[color=#2, mark=#3, mark size=1.2pt, forget plot]
    table[x=G, y=#1, col sep=comma] {diagrams/grad_stab/by_G.csv};
}
\begin{tikzpicture}
\begin{groupplot}[
    group style={group size=4 by 1, horizontal sep=1.0cm},
    scale only axis, width=3.0cm, height=1.7cm,
    tick label style={font=\scriptsize}, label style={font=\footnotesize},
    xlabel style={yshift=0.5ex},
    title style={font=\footnotesize, yshift=-1.2ex},
    axis line style={black!60}, tick style={black!60}, major tick length=2pt,
    grid=major, grid style={black!12, line width=0.3pt},
    every axis plot/.append style={line width=0.9pt, line join=round},
]
\nextgroupplot[title={(a) IIR (\%) vs.\ step}, xlabel={Step}, xmin=0, xmax=234,
    ymin=-0.3, ymax=8,
    legend to name=stablegend, legend columns=3,
    legend style={font=\scriptsize, draw=none, column sep=8pt},
    legend image code/.code={\draw[#1, line width=1.1pt] (0cm,0cm) -- (0.35cm,0cm);}]
\addlegendimage{color=colorGRPO}\addlegendentry{GRPO}
\addlegendimage{color=colorGDPO}\addlegendentry{GDPO}
\addlegendimage{color=colorG3PO}\addlegendentry{G3PO}
\stabstep{iir}{gdpo}{colorGDPO}
\stabstep{iir}{g3po}{colorG3PO}
\nextgroupplot[title={(b) Max advantage vs.\ step}, xlabel={Step}, xmin=0, xmax=234,
    ymin=0, ymax=3.3, ytick={0,1,2,3}]
\addplot[black!45, dashed, line width=0.6pt, forget plot, domain=0:234] {7/sqrt(8)};
\node[font=\tiny, text=black!55, anchor=south east] at (axis cs:232,2.47) {$(G{-}1)/\sqrt{G}$};
\stabstep{maxadv}{grpo}{colorGRPO}
\stabstep{maxadv}{g3po}{colorG3PO}
\nextgroupplot[title={(c) IIR (\%) vs.\ $G$}, xlabel={Group size $G$},
    xtick={2,4,8,16}, xmin=1, xmax=17, ymin=-0.3, ymax=5]
\stabG{iir_gdpo}{colorGDPO}{triangle*}
\stabG{iir_g3po}{colorG3PO}{square*}
\nextgroupplot[title={(d) Max advantage vs.\ $G$}, xlabel={Group size $G$},
    xtick={2,4,8,16}, xmin=1, xmax=17, ymin=0, ymax=4, ytick={0,1,2,3,4}]
\addplot[black!45, dashed, line width=0.6pt, forget plot, domain=2:16, samples=40] {(x-1)/sqrt(x)};
\node[font=\tiny, text=black!55, anchor=east] at (axis cs:11,3.5) {$(G{-}1)/\sqrt{G}$};
\stabG{max_grpo}{colorGRPO}{*}
\stabG{max_g3po}{colorG3PO}{square*}
\end{groupplot}
\node[anchor=south, yshift=0.55cm] at ($(group c2r1.north)!0.5!(group c3r1.north)$) {\pgfplotslegendfromname{stablegend}};
\end{tikzpicture}
    \caption{\chg{Advantage diagnostics on MATH500, computed on the same rollouts. \textbf{(a)} Intra-group inversion rate (IIR: share of correct/incorrect pairs in which the incorrect completion has the larger advantage) over training at $G=8$, for GDPO and G3PO. \textbf{(b)} Maximum advantage over training at $G=8$ for GRPO (main-results reward) and G3PO. \textbf{(c, d)} The same, averaged over training, against $G$. Dashed: GRPO's bound $(G-1)/\sqrt{G}$.}}
    \label{fig:grad_stab}
\end{figure*}
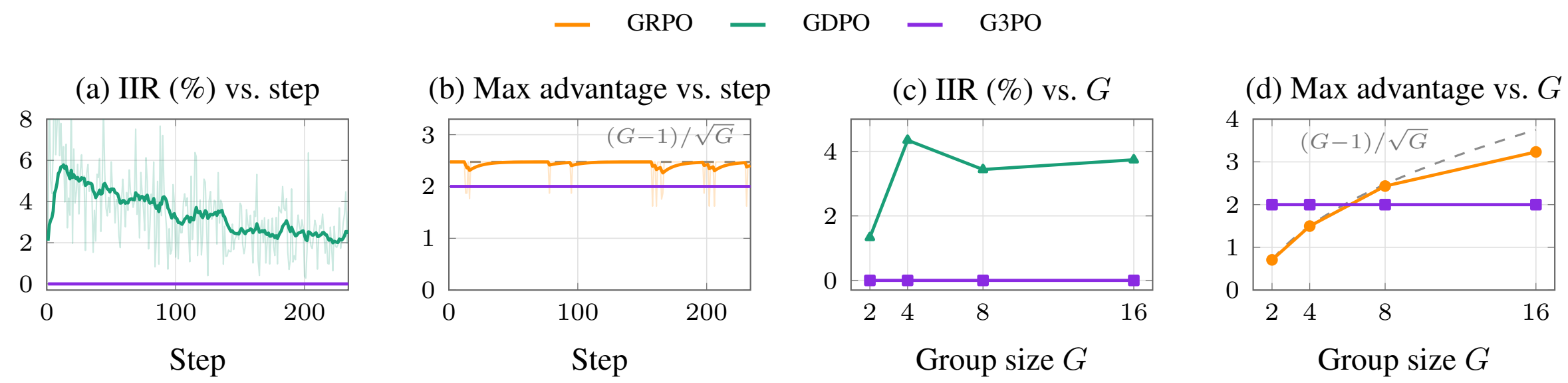

\begin{figure*}[t]
\centering
\definecolor{colorG3PO}{HTML}{8A2BE2}
\definecolor{colorGRPO}{HTML}{FF8C00}
\definecolor{colorGOPO}{HTML}{3B6EA8}
\newcommand{\diagcurves}[2][0.22]{%
  \foreach \m/\c in {grpo/colorGRPO, gopo/colorGOPO, g3po/colorG3PO} {
    \edef\tmp{\noexpand\addplot[color=\c, opacity=#1, line width=0.4pt, forget plot]
      table[x=step, y=raw, col sep=comma] {diagrams/diagnostics/#2_\m.csv};}\tmp
  }
  \addplot[color=colorGRPO] table[x=step, y=ema, col sep=comma] {diagrams/diagnostics/#2_grpo.csv};
  \addplot[color=colorGOPO] table[x=step, y=ema, col sep=comma] {diagrams/diagnostics/#2_gopo.csv};
  \addplot[color=colorG3PO] table[x=step, y=ema, col sep=comma] {diagrams/diagnostics/#2_g3po.csv};
}
\begin{tikzpicture}
\begin{groupplot}[
    group style={group size=3 by 1, horizontal sep=1.25cm},
    scale only axis, width=4.15cm, height=1.8cm,
    xmin=0, xlabel={Training step}, xlabel style={yshift=0.5ex},
    tick label style={font=\scriptsize}, label style={font=\footnotesize},
    title style={font=\footnotesize, yshift=-1ex},
    yticklabel style={/pgf/number format/.cd, fixed, fixed zerofill, precision=2},
    scaled y ticks=false,
    axis line style={black!60}, tick style={black!60}, major tick length=2pt,
    grid=major, grid style={black!12, line width=0.3pt},
    every axis plot/.append style={line width=0.9pt, line join=round},
    legend cell align=left,
    legend image code/.code={\draw[#1, line width=1.2pt] (0cm,0cm) -- (0.4cm,0cm);},
]
\nextgroupplot[title={(a) Gradient norm (pre-clip)}, ymin=0,
    legend to name=diaglegend,
    legend style={font=\scriptsize, draw=none, legend columns=3, column sep=8pt}]
\diagcurves[0.5]{grad_norm}
\legend{GRPO, GOPO, G3PO}
\nextgroupplot[title={(b) KL to reference}, ymin=0,
    yticklabel style={/pgf/number format/.cd, fixed, fixed zerofill, precision=3}]
\diagcurves{kl}
\nextgroupplot[title={(c) Entropy (nats/token)}]
\diagcurves{entropy}
\end{groupplot}
\node[anchor=south, yshift=0.45cm] at (group c2r1.north) {\pgfplotslegendfromname{diaglegend}};
\end{tikzpicture}
\caption{\add{Pre-clip gradient norm, KL divergence to the reference policy and token entropy during training on MATH.}}
\label{fig:diagnostics}
\end{figure*}
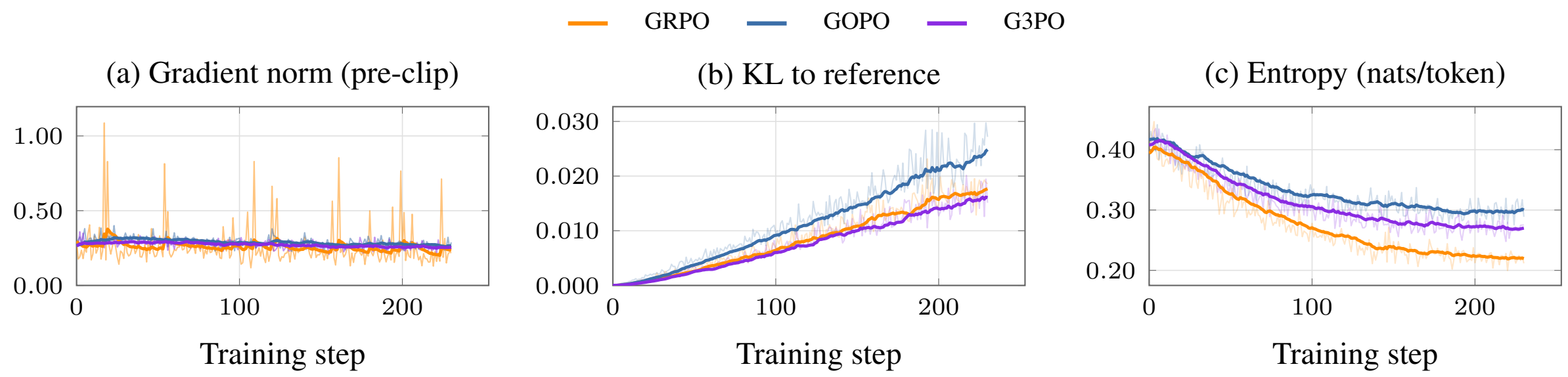

\section{Analyses}

\subsection{Ablation Studies}\label{section/ablations}

We conducted ablation studies to identify the most important components of the G3PO workflow and to assess the viability of our method of combining verifiability with ranking. Table~\ref{ablations} reports results obtained under both GRPO- and GOPO-based advantage formulations across three distinct reward configurations \org{(two for GOPO)}: verifiable outcome reward, Process Reward Model (PRM), and a combination of both.

\org{Under GRPO}, the use of both a verifiable reward and a PRM rarely yields the strongest performance, indicating that this combination, if not used correctly, actually hurts the training signal rather than enhances it. \org{Combined} usage only outperformed its baseline and PRM-only on GSM8K, the easiest mathematical metric. Once the complexity increased in the questions asked, the usefulness of the PRM fluctuated, with the base GRPO, with verifiability only, outputting the better results. 

GOPO results paint a similar story. Of course, its use with a verifiable reward performs very poorly, as it is an algorithm optimised to work alongside a PRM and not involve distinct rewards. 
Overall, these results further reinforce the benefit of correctly integrating verifiability and ranking within a unified reward structure.

\add{\org{Under GRPO, weighting} correctness more ($\alpha=0.8$) helps on the harder benchmarks, while weighting the PRM more ($\alpha=0.2$) hurts \org{on all but GSM8K}. Yet G3PO beats every weighted run on every benchmark evaluated, the strongest (GRPO, $\alpha=0.8$) by \add{1.5} to 4.2 points. GDPO performs on par with GRPO on the unweighted sum (37.6 vs.\ 37.5 on average). The benefit of G3PO therefore comes from the strict priority of the verifier, not from using both signals.}

\subsection{Gradient Stability}
\chg{Figure~\ref{fig:grad_stab} measures two properties of the advantages on MATH500~\cite{Lightman_math500}, over training at $G=8$ and across $G\in\{2,4,8,16\}$. The intra-group inversion rate (IIR) is the share of correct/incorrect pairs in which the incorrect completion receives the larger advantage. G3PO's is zero by construction (Section~\ref{sec:order}). GDPO can invert pairs whose PRM scores differ little, since a narrow in-group PRM spread inflates the normalised PRM term, and it trails G3PO in Table~\ref{ablations}.} \chg{GRPO's maximum advantage grows with $G$ towards $(G-1)/\sqrt{G}$} \chg{while G3PO's stays at $2$, the larger of the two for $G\leq4$. Since a bounded advantage does not imply a smaller gradient (Section~\ref{sec:order}), Figure~\ref{fig:diagnostics} also tracks the update itself.} \add{G3PO's median pre-clip gradient norm is slightly higher than GRPO's (0.27 vs.\ 0.22), as expected, but never exceeds twice its median, whereas GRPO's does on 7\% of steps and peaks at almost five times its median (Figure~\ref{fig:diagnostics}a).} \add{Nor do the gains of Table~\ref{tab:math_benchmarks} come from drifting from the reference policy or collapsing entropy: G3PO ends training with the lowest KL divergence (0.016, against 0.018 for GRPO and 0.024 for GOPO) and an entropy (0.27 nats/token) between GRPO's (0.22) and GOPO's (0.30) (Figure~\ref{fig:diagnostics}b, c).}

\subsection{\texorpdfstring{\add{Robustness to PRM Exploitation}}{Robustness to PRM Exploitation}}\label{sec:hacking}

\begin{table}[t]
\centering
\small
\setlength{\tabcolsep}{5pt}
\renewcommand{\arraystretch}{1.08}
\begin{tabular}{l c S[table-format=2.1] S[table-format=1.2] S[table-format=3.1]}
\toprule
\add{\textbf{Method}} & \add{\textbf{Steps}} & {\add{\textbf{Acc}$\uparrow$}} & {\add{\textbf{PRM}}} & {\add{\textbf{Tokens}}} \\
\midrule
\add{GRPO ($r^v{+}r^p$)} & \add{$T$}  & {\add{76.4}} & {\add{0.65}} & {\add{410.1}} \\
                         & \add{$3T$} & {\add{74.8}} & {\add{0.91}} & {\add{980.7}} \\
\midrule
\rowcolor{g3poTint}
\add{\textbf{G3PO}}      & \add{$T$}  & {\add{78.1}} & {\add{0.64}} & {\add{395.2}} \\
\rowcolor{g3poTint}
                         & \add{$3T$} & {\add{77.6}} & {\add{0.77}} & {\add{740.4}} \\
\bottomrule
\end{tabular}
\caption{\add{Extended training on MATH: accuracy (\%), mean PRM score and mean response length (tokens) after the standard training horizon $T$ and after $3T$.}}
\label{tab:hacking}
\end{table}

{\color{rev}A known risk of PRM-based training is reward hacking, in which the policy learns to raise the PRM score without becoming more accurate. To test G3PO under this pressure, we extend training on MATH to three times the standard horizon $T$ and compare it with GRPO trained on the summed rewards (Table~\ref{tab:hacking}). Both methods show clear signs of PRM exploitation: between $T$ and $3T$, the mean PRM score increases from 0.65 to 0.91 under GRPO and from 0.64 to 0.77 under G3PO, while the mean response length grows by a factor of 2.4 and 1.9, respectively. Accuracy, however, drops by 1.6 points under GRPO and only 0.5 under G3PO. This is consistent with G3PO's design: the PRM only orders completions with the same verifier outcome, so it cannot place an incorrect completion above a correct one. G3PO does not prevent PRM exploitation, but it mitigates the resulting accuracy loss.\par}

\section{Conclusion}

In this paper, we introduced G3PO, a new training framework within the overall Reinforcement Learning paradigm. G3PO intuitively leverages verifiability and preference to maximise reward signals during training of large language models. We show that this new policy optimisation strategy outperforms GRPO and GOPO baselines as well as domain-oriented alternative models across the three areas of Risk of Bias, Medical Reasoning, and Maths. In each of these domains, G3PO consistently remained competitive with, and often surpassed, all compared methods across all benchmarks. Furthermore, our ablations
across reward configurations show that how verifiability and preference are combined matters for
the effectiveness of the training signal. 
Finally, G3PO showed greater training stability than GRPO and, under extended training, lost less accuracy to PRM exploitation, since the PRM can only reorder completions with the same verifier outcome, suggesting a more reliable optimisation behaviour to complement the more consistent evaluation results.

\section*{Limitations}

A central limitation of our approach is its dependence on a reliable verifiability signal during training. Without such a signal, the method reduces to GOPO, removing the main advantage of combining verifiability with preference-based optimisation. Conversely, strong preference signals depend on the availability of a well-trained Process Reward Model (PRM). If the PRM is weak or poorly calibrated, the resulting optimisation signal may be no better than, or in some cases worse than, a simple verifiable reward.

Another important limitation to highlight is the potential for reward hacking that comes with a heavy reliance on PRMs for training. It is a common risk that PRM-based training can be sensitive to exploitable patterns, and such behaviour, although in some cases mitigable, may re-emerge under different settings or at later stages of training.
\add{Extending training to three times the standard horizon (Table~\ref{tab:hacking}) confirms this: G3PO also inflates PRM scores and response length, and only the effect on accuracy is contained.}

{\color{rev}\paragraph{Noisy or unavailable signals.} G3PO trusts the verifier fully. A false negative ranks a correct completion below every completion judged correct, whatever its PRM score, and a false positive ranks an incorrect one above every completion judged incorrect, so verifier errors reach the advantage unattenuated, whereas a weighted sum would let the PRM partly compensate. PRM noise, by contrast, is contained, since the PRM can only reorder completions with the same verifier outcome. When all completions in a group share that outcome, however, G3PO reduces to GOPO on the PRM score for that group, and the update is driven by the PRM alone. Without a PRM, G3PO reduces to a rank-based advantage over the binary verifier reward, in which all correct completions tie.

\paragraph{Scale and scope.} All policies have 7B parameters, and we have not tested whether the gains hold for larger models, for which the 7--8B PRMs we use may be weaker judges than the policy they score. The weighted-sum and GDPO baselines are evaluated on mathematics only, maths results come from a single seed, and the open-ended evaluation (Appendix~\ref{app:rar}) relies on a single LLM judge.\par}







\bibliography{G3PO}

\begin{thebibliography}{50}
\providecommand{\natexlab}[1]{#1}

\bibitem[{Amodei et~al.(2016)Amodei, Olah, Steinhardt, Christiano, Schulman, and Mané}]{amodei_concrete_2016}
Dario Amodei, Chris Olah, Jacob Steinhardt, Paul Christiano, John Schulman, and Dan Mané. 2016.
\newblock \href {https://doi.org/10.48550/arXiv.1606.06565} {Concrete {Problems} in {AI} {Safety}}.
\newblock \emph{arXiv preprint}.
\newblock ArXiv:1606.06565 [cs].

\bibitem[{{Anthropic}(2026)}]{anthropic2026opus5}
{Anthropic}. 2026.
\newblock {Claude Opus 5}.
\newblock \url{https://www.anthropic.com/claude}.

\bibitem[{Arora et~al.(2025)Arora, Wei, Hicks, Bowman, Qui{\~n}onero-Candela, Tsimpourlas, Sharman, Shah, Vallone, Beutel, Heidecke, and Singhal}]{arora2025healthbench}
Rahul~K. Arora, Jason Wei, Rebecca~Soskin Hicks, Preston Bowman, Joaquin Qui{\~n}onero-Candela, Foivos Tsimpourlas, Michael Sharman, Meghan Shah, Andrea Vallone, Alex Beutel, Johannes Heidecke, and Karan Singhal. 2025.
\newblock {HealthBench}: Evaluating large language models towards improved human health.
\newblock \emph{arXiv preprint arXiv:2505.08775}.

\bibitem[{Chen et~al.(2026)Chen, Zheng, Zhang, Cui, Cui, Ye, Lin, Liu, Zhu, and Wang}]{NFT}
Huayu Chen, Kaiwen Zheng, Qinsheng Zhang, Ganqu Cui, Yin Cui, Haotian Ye, Tsung-Yi Lin, Ming-Yu Liu, Jun Zhu, and Haoxiang Wang. 2026.
\newblock \href {https://openreview.net/forum?id=ujBrsQm6Zu} {{NFT}: Bridging supervised learning and reinforcement learning in math reasoning}.
\newblock In \emph{The Fourteenth International Conference on Learning Representations}.

\bibitem[{Chen et~al.(2025)Chen, Cai, Ji, Wang, Liu, Wang, and Wang}]{chen2024huatuogpt}
Junying Chen, Zhenyang Cai, Ke~Ji, Xidong Wang, Wanlong Liu, Rongsheng Wang, and Benyou Wang. 2025.
\newblock \href {https://doi.org/10.18653/v1/2025.findings-acl.751} {Towards medical complex reasoning with {LLM}s through medical verifiable problems}.
\newblock In \emph{Findings of the Association for Computational Linguistics: ACL 2025}, pages 14552--14573, Vienna, Austria. Association for Computational Linguistics.

\bibitem[{Choi et~al.(2026)Choi, Saha, Kim, Agarwal, and Dwivedi}]{choi_gopo_2026}
Kyuseong Choi, Dwaipayan Saha, Woojeong Kim, Anish Agarwal, and Raaz Dwivedi. 2026.
\newblock \href {https://doi.org/10.48550/arXiv.2602.03876} {{GOPO}: {Policy} {Optimization} using {Ranked} {Rewards}}.
\newblock \emph{arXiv preprint}.
\newblock ArXiv:2602.03876 [cs.LG].

\bibitem[{Cobbe et~al.(2021)Cobbe, Kosaraju, Bavarian, Chen, Jun, Kaiser, Plappert, Tworek, Hilton, Nakano, Hesse, and Schulman}]{cobbe_gsm8k}
Karl Cobbe, Vineet Kosaraju, Mohammad Bavarian, Mark Chen, Heewoo Jun, Lukasz Kaiser, Matthias Plappert, Jerry Tworek, Jacob Hilton, Reiichiro Nakano, Christopher Hesse, and John Schulman. 2021.
\newblock Training verifiers to solve math word problems.
\newblock \emph{arXiv preprint arXiv:2110.14168}.

\bibitem[{Cui et~al.(2026)Cui, Yuan, Wang, Wang, Zhang, Chen, Li, He, Fan, Yu, Xu, Chen, Yuan, Chen, Zhang, Lv, Wang, Yao, Han, Peng, Cheng, Liu, Sun, Zhou, and Ding}]{Eurus-2-7B}
Ganqu Cui, Lifan Yuan, Zefan Wang, Hanbin Wang, Yuchen Zhang, Jiacheng Chen, Wendi Li, Bingxiang He, Yuchen Fan, Tianyu Yu, Qixin Xu, Weize Chen, Jiarui Yuan, Huayu Chen, Kaiyan Zhang, Xingtai Lv, Shuo Wang, Yuan Yao, Xu~Han, and 6 others. 2026.
\newblock \href {https://openreview.net/forum?id=9SkkifLopZ} {Process reinforcement through implicit rewards}.
\newblock \emph{Transactions on Machine Learning Research}.

\bibitem[{Gunjal et~al.(2026)Gunjal, Wang, Lau, Nath, He, Liu, and Hendryx}]{gunjal2026rubrics}
Anisha Gunjal, Anthony Wang, Elaine Lau, Vaskar Nath, Yunzhong He, Bing Liu, and Sean Hendryx. 2026.
\newblock Rubrics as rewards: Reinforcement learning beyond verifiable domains.
\newblock In \emph{The Fourteenth International Conference on Learning Representations}.

\bibitem[{Guo et~al.(2025)Guo, Yang, Zhang, Song, Wang, Zhu, Xu, Zhang, Ma, Bi, Zhang, Yu, Wu, Wu, Gou, Shao, Li, Gao, Liu, Xue, Wang, Wu, Feng, Lu, Zhao, Deng, Ruan, Dai, Chen, Ji, Li, Lin, Dai, Luo, Hao, Chen, Li, Zhang, Xu, Ding, Gao, Qu, Li, Guo, Li, Chen, Yuan, Tu, Qiu, Li, Cai, Ni, Liang, Chen, Dong, Hu, You, Gao, Guan, Huang, Yu, Wang, Zhang, Zhao, Wang, Zhang, Xu, Xia, Zhang, Zhang, Tang, Zhou, Li, Wang, Li, Tian, Huang, Zhang, Wang, Chen, Du, Ge, Zhang, Pan, Wang, Chen, Jin, Chen, Lu, Zhou, Chen, Ye, Wang, Yu, Zhou, Pan, Li, Zhou, Wu, Yun, Pei, Sun, Wang, Zeng, Liu, Liang, Gao, Yu, Zhang, Xiao, An, Liu, Wang, Chen, Nie, Cheng, Liu, Xie, Liu, Yang, Li, Su, Lin, Li, Jin, Shen, Chen, Sun, Wang, Song, Zhou, Wang, Shan, Li, Wang, Wei, Zhang, Xu, Li, Zhao, Sun, Wang, Yu, Zhang, Shi, Xiong, He, Piao, Wang, Tan, Ma, Liu, Guo, Ou, Wang, Gong, Zou, He, Xiong, Luo, You, Liu, Zhou, Zhu, Huang, Li, Zheng, Zhu, Ma, Tang, Zha, Yan, Ren, Ren, Sha, Fu, Xu, Xie, Zhang, Hao, Ma, Yan, Wu, Gu, Zhu, Liu, Li, Xie, Song,
  Pan, Huang, Xu, Zhang, and Zhang}]{guo_deepseek-r1_2025}
Daya Guo, Dejian Yang, Haowei Zhang, Junxiao Song, Peiyi Wang, Qihao Zhu, Runxin Xu, Ruoyu Zhang, Shirong Ma, Xiao Bi, Xiaokang Zhang, Xingkai Yu, Yu~Wu, Z.~F. Wu, Zhibin Gou, Zhihong Shao, Zhuoshu Li, Ziyi Gao, Aixin Liu, and 175 others. 2025.
\newblock \href {https://doi.org/10.1038/s41586-025-09422-z} {{DeepSeek}-{R1} incentivizes reasoning in {LLMs} through reinforcement learning}.
\newblock \emph{Nature}, 645(8081):633--638.

\bibitem[{Hasselt et~al.(2018)Hasselt, Doron, Strub, Hessel, Sonnerat, and Modayil}]{hasselt_deep_2018}
Hado~van Hasselt, Yotam Doron, Florian Strub, Matteo Hessel, Nicolas Sonnerat, and Joseph Modayil. 2018.
\newblock \href {https://doi.org/10.48550/arXiv.1812.02648} {Deep {Reinforcement} {Learning} and the {Deadly} {Triad}}.
\newblock \emph{arXiv preprint}.
\newblock ArXiv:1812.02648.

\bibitem[{He et~al.(2024)He, Luo, Bai, Hu, Thai, Shen, Hu, Han, Huang, Zhang, Liu, Qi, Liu, and Sun}]{olympiadbench}
Chaoqun He, Renjie Luo, Yuzhuo Bai, Shengding Hu, Zhen Thai, Junhao Shen, Jinyi Hu, Xu~Han, Yujie Huang, Yuxiang Zhang, Jie Liu, Lei Qi, Zhiyuan Liu, and Maosong Sun. 2024.
\newblock \href {https://doi.org/10.18653/v1/2024.acl-long.211} {{O}lympiad{B}ench: A challenging benchmark for promoting {AGI} with olympiad-level bilingual multimodal scientific problems}.
\newblock In \emph{Proceedings of the 62nd Annual Meeting of the Association for Computational Linguistics (Volume 1: Long Papers)}, pages 3828--3850, Bangkok, Thailand. Association for Computational Linguistics.

\bibitem[{Hendrycks et~al.(2021)Hendrycks, Burns, Kadavath, Arora, Basart, Tang, Song, and Steinhardt}]{hendrycks2021measuring_mathdataset}
Dan Hendrycks, Collin Burns, Saurav Kadavath, Akul Arora, Steven Basart, Eric Tang, Dawn Song, and Jacob Steinhardt. 2021.
\newblock \href {https://openreview.net/forum?id=7Bywt2mQsCe} {Measuring mathematical problem solving with the {MATH} dataset}.
\newblock In \emph{Thirty-fifth Conference on Neural Information Processing Systems Datasets and Benchmarks Track (Round 2)}.

\bibitem[{Higgins et~al.(2024)Higgins, Thomas, Chandler, Cumpston, Li, Page, and Welch}]{higgins2024}
J.~P.~T. Higgins, J.~Thomas, J.~Chandler, M.~Cumpston, T.~Li, M.~J. Page, and V.~A. Welch, editors. 2024.
\newblock \href {https://training.cochrane.org/handbook} {\emph{Cochrane Handbook for Systematic Reviews of Interventions}}, 6.5 edition.
\newblock Cochrane, London, U.K.

\bibitem[{Huang et~al.(2026)Huang, Wu, Liu, Tang, and Zhou}]{m23k}
Xiaoke Huang, Juncheng Wu, Hui Liu, Xianfeng Tang, and Yuyin Zhou. 2026.
\newblock \href {https://proceedings.mlr.press/v297/huang26a.html} {m1: Unleash the potential of test-time scaling for medical reasoning with large language models}.
\newblock In \emph{Proceedings of the Fifth Machine Learning for Health Symposium}, volume 297 of \emph{Proceedings of Machine Learning Research}, pages 369--383. PMLR.

\bibitem[{{Hugging Face}(2025)}]{openr1}
{Hugging Face}. 2025.
\newblock \href {https://github.com/huggingface/open-r1} {Open {R}1: A fully open reproduction of {D}eep{S}eek-{R}1}.

\bibitem[{Jin et~al.(2021)Jin, Pan, Oufattole, Weng, Fang, and Szolovits}]{medqa}
Di~Jin, Eileen Pan, Nassim Oufattole, Wei-Hung Weng, Hanyi Fang, and Peter Szolovits. 2021.
\newblock What disease does this patient have? a large-scale open domain question answering dataset from medical exams.
\newblock \emph{Applied Sciences}, 11(14):6421.

\bibitem[{Kwon et~al.(2023)Kwon, Li, Zhuang, Sheng, Zheng, Yu, Gonzalez, Zhang, and Stoica}]{vllm}
Woosuk Kwon, Zhuohan Li, Siyuan Zhuang, Ying Sheng, Lianmin Zheng, Cody~Hao Yu, Joseph Gonzalez, Hao Zhang, and Ion Stoica. 2023.
\newblock Efficient memory management for large language model serving with pagedattention.
\newblock In \emph{Proceedings of the 29th symposium on operating systems principles}, pages 611--626.

\bibitem[{Lambert et~al.(2025)Lambert, Morrison, Pyatkin, Huang, Ivison, Brahman, Miranda, Liu, Dziri, Lyu, Gu, Malik, Graf, Hwang, Yang, Bras, Tafjord, Wilhelm, Soldaini, Smith, Wang, Dasigi, and Hajishirzi}]{lambert2024tulu}
Nathan Lambert, Jacob Morrison, Valentina Pyatkin, Shengyi Huang, Hamish Ivison, Faeze Brahman, Lester James~Validad Miranda, Alisa Liu, Nouha Dziri, Xinxi Lyu, Yuling Gu, Saumya Malik, Victoria Graf, Jena~D. Hwang, Jiangjiang Yang, Ronan~Le Bras, Oyvind Tafjord, Christopher Wilhelm, Luca Soldaini, and 4 others. 2025.
\newblock \href {https://openreview.net/forum?id=i1uGbfHHpH} {Tulu 3: Pushing frontiers in open language model post-training}.
\newblock In \emph{Second Conference on Language Modeling}.

\bibitem[{Lightman et~al.(2024)Lightman, Kosaraju, Burda, Edwards, Baker, Lee, Leike, Schulman, Sutskever, and Cobbe}]{Lightman_math500}
Hunter Lightman, Vineet Kosaraju, Yuri Burda, Harrison Edwards, Bowen Baker, Teddy Lee, Jan Leike, John Schulman, Ilya Sutskever, and Karl Cobbe. 2024.
\newblock \href {https://openreview.net/forum?id=v8L0pN6EOi} {Let's verify step by step}.
\newblock In \emph{The Twelfth International Conference on Learning Representations}.

\bibitem[{Liu et~al.(2025{\natexlab{a}})Liu, Wang, Pan, Wan, Dai, Lin, Bai, Rueckert, and Arcucci}]{alphamed}
Che Liu, Haozhe Wang, Jiazhen Pan, Zhongwei Wan, Yong Dai, Fangzhen Lin, Wenjia Bai, Daniel Rueckert, and Rossella Arcucci. 2025{\natexlab{a}}.
\newblock Beyond distillation: Pushing the limits of medical llm reasoning with minimalist rule-based rl.
\newblock \emph{arXiv preprint arXiv:2505.17952}.

\bibitem[{Liu et~al.(2026)Liu, Dong, Lu, Diao, Belcak, Liu, Chen, Yin, Wang, Cheng, Choi, Kautz, and Molchanov}]{liu2026gdpo}
Shih-Yang Liu, Xin Dong, Ximing Lu, Shizhe Diao, Peter Belcak, Mingjie Liu, Min-Hung Chen, Hongxu Yin, Yu-Chiang~Frank Wang, Kwang-Ting Cheng, Yejin Choi, Jan Kautz, and Pavlo Molchanov. 2026.
\newblock \href {https://openreview.net/forum?id=Y1rHxA58If} {{GDPO}: Group reward-decoupled normalization policy optimization for multi-reward {RL} optimization}.
\newblock In \emph{Forty-third International Conference on Machine Learning}.

\bibitem[{Liu et~al.(2025{\natexlab{b}})Liu, Chen, Li, Qi, Pang, Du, Lee, and Lin}]{OaT-Zero}
Zichen Liu, Changyu Chen, Wenjun Li, Penghui Qi, Tianyu Pang, Chao Du, Wee~Sun Lee, and Min Lin. 2025{\natexlab{b}}.
\newblock Understanding {R}1-{Z}ero-{L}ike training: A critical perspective.
\newblock In \emph{Conference on Language Modeling (COLM)}.

\bibitem[{Loshchilov and Hutter(2019)}]{adamw}
Ilya Loshchilov and Frank Hutter. 2019.
\newblock \href {https://openreview.net/forum?id=Bkg6RiCqY7} {Decoupled weight decay regularization}.
\newblock In \emph{International Conference on Learning Representations}.

\bibitem[{math{-}ai(2024)}]{amc23}
math{-}ai. 2024.
\newblock {AMC}23: A mathematical reasoning benchmark from the 2023 american mathematics competitions.
\newblock \url{https://huggingface.co/datasets/math-ai/amc23}.

\bibitem[{Ouyang et~al.(2022)Ouyang, Wu, Jiang, Almeida, Wainwright, Mishkin, Zhang, Agarwal, Slama, Ray, Schulman, Hilton, Kelton, Miller, Simens, Askell, Welinder, Christiano, Leike, and Lowe}]{Ouyang_RLHF}
Long Ouyang, Jeff Wu, Xu~Jiang, Diogo Almeida, Carroll~L. Wainwright, Pamela Mishkin, Chong Zhang, Sandhini Agarwal, Katarina Slama, Alex Ray, John Schulman, Jacob Hilton, Fraser Kelton, Luke Miller, Maddie Simens, Amanda Askell, Peter Welinder, Paul Christiano, Jan Leike, and Ryan Lowe. 2022.
\newblock Training language models to follow instructions with human feedback.
\newblock In \emph{Proceedings of the 36th International Conference on Neural Information Processing Systems}, NIPS '22, Red Hook, NY, USA. Curran Associates Inc.

\bibitem[{Pal et~al.(2022)Pal, Umapathi, and Sankarasubbu}]{medmcqa}
Ankit Pal, Logesh~Kumar Umapathi, and Malaikannan Sankarasubbu. 2022.
\newblock \href {https://proceedings.mlr.press/v174/pal22a.html} {Med{MCQA}: A large-scale multi-subject multi-choice dataset for medical domain question answering}.
\newblock In \emph{Proceedings of the Conference on Health, Inference, and Learning}, volume 174 of \emph{Proceedings of Machine Learning Research}, pages 248--260. PMLR.

\bibitem[{Pronesti et~al.(2026)Pronesti, Belz, and Hou}]{pronesti2026beyond}
Massimiliano Pronesti, Anya Belz, and Yufang Hou. 2026.
\newblock \href {https://doi.org/10.18653/v1/2026.findings-acl.1611} {Beyond outcome verification: Verifiable process reward models for structured reasoning}.
\newblock In \emph{Findings of the {A}ssociation for {C}omputational {L}inguistics: {ACL} 2026}, pages 32187--32202, San Diego, California, United States. Association for Computational Linguistics.

\bibitem[{Pronesti et~al.(2025{\natexlab{a}})Pronesti, Bettencourt-Silva, Flanagan, Pascale, Redmond, Belz, and Hou}]{pronesti_query-driven_2025}
Massimiliano Pronesti, Joao~H Bettencourt-Silva, Paul Flanagan, Alessandra Pascale, Oisín Redmond, Anya Belz, and Yufang Hou. 2025{\natexlab{a}}.
\newblock \href {https://doi.org/10.18653/v1/2025.acl-long.1359} {Query-driven {Document}-level {Scientific} {Evidence} {Extraction} from {Biomedical} {Studies}}.
\newblock In \emph{Proceedings of the 63rd {Annual} {Meeting} of the {Association} for {Computational} {Linguistics} ({Volume} 1: {Long} {Papers})}, pages 28034--28051, Vienna, Austria. Association for Computational Linguistics.

\bibitem[{Pronesti et~al.(2025{\natexlab{b}})Pronesti, Lorandi, Flanagan, Redmond, Belz, and Hou}]{pronesti_enhancing_2025}
Massimiliano Pronesti, Michela Lorandi, Paul Flanagan, Oisín Redmond, Anya Belz, and Yufang Hou. 2025{\natexlab{b}}.
\newblock \href {https://doi.org/10.18653/v1/2025.emnlp-main.1544} {Enhancing {Study}-{Level} {Inference} from {Clinical} {Trial} {Papers} via {Reinforcement} {Learning}-{Based} {Numeric} {Reasoning}}.
\newblock In \emph{Proceedings of the 2025 {Conference} on {Empirical} {Methods} in {Natural} {Language} {Processing}}, pages 30357--30373, Suzhou, China. Association for Computational Linguistics.

\bibitem[{{ROBINS-I}(2025)}]{robins2025}
{ROBINS-I}. 2025.
\newblock \href {https://www.riskofbias.info/welcome/robins-i-v2} {The risk of bias in non-randomized studies – of interventions, version 2 ({ROBINS-I V2}) assessment tool (for follow-up studies)}.
\newblock Assessment tool, University of Bristol, Bristol, U.K.

\bibitem[{Schulman(2020)}]{schulman}
John Schulman. 2020.
\newblock \href {http://joschu.net/blog/kl-approx.html} {Approximating {KL} {Divergence}}.

\bibitem[{Schulman et~al.(2017)Schulman, Wolski, Dhariwal, Radford, and Klimov}]{schulman_proximal_2017}
John Schulman, Filip Wolski, Prafulla Dhariwal, Alec Radford, and Oleg Klimov. 2017.
\newblock \href {https://doi.org/10.48550/arXiv.1707.06347} {Proximal {Policy} {Optimization} {Algorithms}}.
\newblock \emph{arXiv preprint}.
\newblock ArXiv:1707.06347.

\bibitem[{Shao et~al.(2024)Shao, Wang, Zhu, Xu, Song, Bi, Zhang, Zhang, Li, Wu et~al.}]{shao_deepseekmath_2024}
Zhihong Shao, Peiyi Wang, Qihao Zhu, Runxin Xu, Junxiao Song, Xiao Bi, Haowei Zhang, Mingchuan Zhang, YK~Li, Yang Wu, et~al. 2024.
\newblock Deepseekmath: Pushing the limits of mathematical reasoning in open language models.
\newblock \emph{arXiv preprint arXiv:2402.03300}.

\bibitem[{Sterne et~al.(2019)Sterne, Savovi{\'c}, Page, Elbers, Blencowe, Boutron, Cates, Cheng, Corbett, Eldridge, Emberson, Hern{\'a}n, Hopewell, Hr{\'o}bjartsson, Junqueira, J{\"u}ni, Kirkham, Lasserson, Li, McAleenan, Reeves, Shepperd, Shrier, Stewart, Tilling, White, Whiting, and Higgins}]{sterne2019}
Jonathan A.~C. Sterne, Jelena Savovi{\'c}, Matthew~J. Page, Roy~G. Elbers, Natalie~S. Blencowe, Isabelle Boutron, Christopher~J. Cates, Hung-Yuan Cheng, Mark~S. Corbett, Sandra~M. Eldridge, Jonathan~R. Emberson, Miguel~A. Hern{\'a}n, Sally Hopewell, Asbj{\"o}rn Hr{\'o}bjartsson, Daniela~R. Junqueira, Peter J{\"u}ni, Jamie~J. Kirkham, Toby Lasserson, Tianjing Li, and 9 others. 2019.
\newblock \href {https://doi.org/10.1136/bmj.l4898} {Rob 2: a revised tool for assessing risk of bias in randomised trials}.
\newblock \emph{BMJ}, 366:l4898.

\bibitem[{Su et~al.(2026)Su, Yu, Song, Li, Mi, Tu, Zhang, and Yu}]{su2025crossing}
Yi~Su, Dian Yu, Linfeng Song, Juntao Li, Haitao Mi, Zhaopeng Tu, Min Zhang, and Dong Yu. 2026.
\newblock \href {https://doi.org/10.18653/v1/2026.acl-long.178} {Crossing the reward bridge: Expanding reinforcement learning with verifiable rewards across diverse domains}.
\newblock In \emph{Proceedings of the 64th Annual Meeting of the {A}ssociation for {C}omputational {L}inguistics (Volume 1: Long Papers)}, pages 3872--3892, San Diego, California, United States. Association for Computational Linguistics.

\bibitem[{Wang et~al.(2024{\natexlab{a}})Wang, Li, Shao, Xu, Dai, Li, Chen, Wu, and Sui}]{wang_math-shepherd_2024}
Peiyi Wang, Lei Li, Zhihong Shao, Runxin Xu, Damai Dai, Yifei Li, Deli Chen, Yu~Wu, and Zhifang Sui. 2024{\natexlab{a}}.
\newblock \href {https://doi.org/10.18653/v1/2024.acl-long.510} {Math-{Shepherd}: {Verify} and {Reinforce} {LLMs} {Step}-by-step without {Human} {Annotations}}.
\newblock In \emph{Proceedings of the 62nd {Annual} {Meeting} of the {Association} for {Computational} {Linguistics} ({Volume} 1: {Long} {Papers})}, pages 9426--9439, Bangkok, Thailand. Association for Computational Linguistics.

\bibitem[{Wang et~al.(2024{\natexlab{b}})Wang, Ma, Zhang, Ni, Chandra, Guo, Ren, Arulraj, He, Jiang et~al.}]{wang2024mmlu}
Yubo Wang, Xueguang Ma, Ge~Zhang, Yuansheng Ni, Abhranil Chandra, Shiguang Guo, Weiming Ren, Aaran Arulraj, Xuan He, Ziyan Jiang, et~al. 2024{\natexlab{b}}.
\newblock {MMLU}-{P}ro: A more robust and challenging multi-task language understanding benchmark.
\newblock \emph{Advances in Neural Information Processing Systems}, 37:95266--95290.

\bibitem[{Wang et~al.(2025)Wang, Gao, and Ke}]{wang2025masked}
Zhen Wang, Zhifeng Gao, and Guolin Ke. 2025.
\newblock Masked-and-reordered self-supervision for reinforcement learning from verifiable rewards.
\newblock \emph{arXiv preprint arXiv:2511.17473}.

\bibitem[{Wen et~al.(2026)Wen, Liu, Zheng, Ye, Wu, Wang, Xu, Liang, Li, Miao, Bian, and Yang}]{wen2025reinforcement}
Xumeng Wen, Zihan Liu, Shun Zheng, Shengyu Ye, Zhirong Wu, Yang Wang, Zhijian Xu, Xiao Liang, Junjie Li, Ziming Miao, Jiang Bian, and Mao Yang. 2026.
\newblock \href {https://openreview.net/forum?id=jGbRWwIidy} {Reinforcement learning with verifiable rewards implicitly incentivizes correct reasoning in base {LLM}s}.
\newblock In \emph{The Fourteenth International Conference on Learning Representations}.

\bibitem[{Wu et~al.(2026)Wu, Deng, Li, Liu, Mi, Peng, Xu, Liu, Cho, Choi, Cao, Ren, Li, Li, and Zhou}]{medreason}
Juncheng Wu, Wenlong Deng, Xingxuan Li, Sheng Liu, Taomian Mi, Yifan Peng, Ziyang Xu, Yi~Liu, Hyunjin Cho, Chang-In Choi, Yihan Cao, Hui Ren, Xiang Li, Xiaoxiao Li, and Yuyin Zhou. 2026.
\newblock \href {https://aclanthology.org/2026.groundlm-1.1/} {{M}ed{R}eason: Eliciting factual medical reasoning steps in {LLM}s via knowledge graphs}.
\newblock In \emph{Proceedings of the 1st Workshop on Grounding Language Models: Learning Faithfully and Efficiently ({G}round{LM} 2026)}. Association for Computational Linguistics.

\bibitem[{Yang et~al.(2025)Yang, Yang, Zhang, Hui, Zheng, Yu, Li, Liu, Huang, Wei, Lin, Yang, Tu, Zhang, Yang, Yang, Zhou, Lin, Dang, Lu, Bao, Yang, Yu, Li, Xue, Zhang, Zhu, Men, Lin, Li, Tang, Xia, Ren, Ren, Fan, Su, Zhang, Wan, Liu, Cui, Zhang, and Qiu}]{qwen2.5}
An~Yang, Baosong Yang, Beichen Zhang, Binyuan Hui, Bo~Zheng, Bowen Yu, Chengyuan Li, Dayiheng Liu, Fei Huang, Haoran Wei, Huan Lin, Jian Yang, Jianhong Tu, Jianwei Zhang, Jianxin Yang, Jiaxi Yang, Jingren Zhou, Junyang Lin, Kai Dang, and 23 others. 2025.
\newblock \href {https://arxiv.org/abs/2412.15115} {Qwen2.5 technical report}.
\newblock \emph{Preprint}, arXiv:2412.15115.

\bibitem[{Yang et~al.(2024)Yang, Zhang, Hui, Gao, Yu, Li, Liu, Tu, Zhou, Lin, Lu, Xue, Lin, Liu, Ren, and Zhang}]{qwen2.5Math}
An~Yang, Beichen Zhang, Binyuan Hui, Bofei Gao, Bowen Yu, Chengpeng Li, Dayiheng Liu, Jianhong Tu, Jingren Zhou, Junyang Lin, Keming Lu, Mingfeng Xue, Runji Lin, Tianyu Liu, Xingzhang Ren, and Zhenru Zhang. 2024.
\newblock Qwen2.5-math technical report: Toward mathematical expert model via self-improvement.
\newblock \emph{arXiv preprint arXiv:2409.12122}.

\bibitem[{Yu et~al.(2025)Yu, Zhang, Zhu, Yuan, Zuo, YuYue, Dai, Fan, Liu, Liu, Liu, Liu, Lin, Lin, Ma, Sheng, Tong, Zhang, Zhang, Zhang, Zhang, Zhu, Zhu, Chen, Chen, Wang, Yu, Song, Wei, Zhou, Liu, Ma, Zhang, Yan, Wu, and Wang}]{yu2025dapo}
Qiying Yu, Zheng Zhang, Ruofei Zhu, Yufeng Yuan, Xiaochen Zuo, YuYue, Weinan Dai, Tiantian Fan, Gaohong Liu, Juncai Liu, LingJun Liu, Xin Liu, Haibin Lin, Zhiqi Lin, Bole Ma, Guangming Sheng, Yuxuan Tong, Chi Zhang, Mofan Zhang, and 17 others. 2025.
\newblock \href {https://openreview.net/forum?id=2a36EMSSTp} {{DAPO}: An open-source {LLM} reinforcement learning system at scale}.
\newblock In \emph{The Thirty-ninth Annual Conference on Neural Information Processing Systems}.

\bibitem[{Yue et~al.(2026)Yue, Chen, Lu, Zhao, Wang, Yue, Song, and Huang}]{yue2026does}
Yang Yue, Zhiqi Chen, Rui Lu, Andrew Zhao, Zhaokai Wang, Yang Yue, Shiji Song, and Gao Huang. 2026.
\newblock \href {https://openreview.net/forum?id=4OsgYD7em5} {Does reinforcement learning really incentivize reasoning capacity in {LLM}s beyond the base model?}
\newblock In \emph{The Thirty-ninth Annual Conference on Neural Information Processing Systems}.

\bibitem[{Yun et~al.(2025)Yun, Sohn, Park, Kim, Tang, Shao, Koo, Minhyeok, Chen, Gerstein et~al.}]{medprm}
Jaehoon Yun, Jiwoong Sohn, Jungwoo Park, Hyunjae Kim, Xiangru Tang, Daniel Shao, Yong~Hoe Koo, Ko~Minhyeok, Qingyu Chen, Mark Gerstein, et~al. 2025.
\newblock Med-prm: Medical reasoning models with stepwise, guideline-verified process rewards.
\newblock In \emph{Proceedings of the 2025 Conference on Empirical Methods in Natural Language Processing}, pages 16565--16582.

\bibitem[{Zhang and Math-AI(2024)}]{aime24}
Yifan Zhang and Team Math-AI. 2024.
\newblock American invitational mathematics examination ({AIME}) 2024.

\bibitem[{Zhang and Math-AI(2025)}]{aime25}
Yifan Zhang and Team Math-AI. 2025.
\newblock American invitational mathematics examination ({AIME}) 2025.

\bibitem[{Zhang and Math-AI(2026)}]{aime26}
Yifan Zhang and Team Math-AI. 2026.
\newblock American invitational mathematics examination ({AIME}) 2026.

\bibitem[{Zhang et~al.(2025)Zhang, Zheng, Wu, Zhang, Lin, Yu, Liu, Zhou, and Lin}]{zhang_lessons_2025}
Zhenru Zhang, Chujie Zheng, Yangzhen Wu, Beichen Zhang, Runji Lin, Bowen Yu, Dayiheng Liu, Jingren Zhou, and Junyang Lin. 2025.
\newblock \href {https://doi.org/10.18653/v1/2025.findings-acl.547} {The {Lessons} of {Developing} {Process} {Reward} {Models} in {Mathematical} {Reasoning}}.
\newblock In \emph{Findings of the {Association} for {Computational} {Linguistics}: {ACL} 2025}, pages 10495--10516, Vienna, Austria. Association for Computational Linguistics.

\end{thebibliography}

\appendix
\makeatletter\setlength{\@fptop}{0pt}\setlength{\@fpsep}{14pt plus 2pt}\makeatother 

\section{Training Details}\label{app:hyperparams} We executed all the experiments on our own distributed cluster. We trained two 7B models: Qwen2.5-7B-Instruct~\cite{qwen2.5}, used for Risk of Bias Assessment \& Medical Reasoning, and Qwen2.5-7B-Math~\cite{qwen2.5Math} used for mathematical tasks. Both models were trained for 1 epoch, with 8 generated completions per question and learning rates ranging from $5.0 \times 10^{-7}$ for medical tasks and Maths, and $3.0 \times 10^{-6}$ for RoB Assessment with the AdamW optimiser~\cite{adamw}. The maximum generated steps were capped at $2048$ for all domains.

For the RL setups, we adopt the GRPO~\cite{shao_deepseekmath_2024} and GOPO~\cite{choi_gopo_2026} algorithms, as well as our own G3PO method. Both training protocols leverage gradient accumulation with 8 accumulation steps. All experiments are conducted using the Open-R1
framework~\cite{openr1} on 8 NVIDIA A100 GPUs, each equipped with 80GB of memory. Models have been served for inference with the vLLM framework~\cite{vllm}. Detailed training hyperparameters are listed in Table~\ref{tab:hyp}.

\begin{table}[!h]
\centering
\footnotesize
\setlength{\tabcolsep}{3pt}
\begin{tabular*}{\columnwidth}{@{\extracolsep{\fill}} l ccc}
\toprule
\textbf{Hyperparameter} & \textbf{RoB} & \textbf{Medical} & \textbf{Maths} \\
\midrule
\grouprow{4}{Domain-specific} \\
\quad Learning rate               & 3e-6 & 5e-7 & 5e-7 \\
\quad Per-device batch size       & 1 & 8 & 8 \\
\quad Grad.\ accumulation       & 8 & 4 & 4 \\
\midrule
\grouprow{4}{Shared across domains} \\
\quad Group size $G$            & \multicolumn{3}{c}{8} \\
\quad Epochs                      & \multicolumn{3}{c}{1} \\
\quad Max.\ completion length     & \multicolumn{3}{c}{2048} \\
\quad KL coefficient $\beta$      & \multicolumn{3}{c}{0.001} \\
\quad Optimiser                   & \multicolumn{3}{c}{AdamW (8-bit)} \\
\quad LR schedule                 & \multicolumn{3}{c}{cosine} \\
\quad Warm-up ratio               & \multicolumn{3}{c}{0.03} \\
\quad Precision                   & \multicolumn{3}{c}{bfloat16} \\
\quad vLLM GPU memory          & \multicolumn{3}{c}{0.5} \\
\bottomrule
\end{tabular*}
\caption{Training and inference hyperparameters.}
\label{tab:hyp}
\end{table}

\section{Training Dynamics}
\begin{figure}[t]
\centering
\begin{tikzpicture}
\definecolor{colorG3PO}{HTML}{8A2BE2}
\definecolor{colorGRPO}{HTML}{FF8C00}
\definecolor{colorDAPO}{HTML}{50C080}
\begin{axis}[
    width=\columnwidth,
    height=0.52\columnwidth,
    xlabel={Training step},
    ylabel={Outcome reward},
    xmin=0, xmax=230,
    ymin=0.40, ymax=0.90,
    xtick={0,50,100,150,200},
    ytick={0.4,0.5,0.6,0.7,0.8,0.9},
    yticklabel style={/pgf/number format/.cd, fixed, fixed zerofill, precision=1},
    tick label style={font=\scriptsize},
    label style={font=\footnotesize},
    axis line style={black!60},
    tick style={black!60},
    major tick length=2pt,
    grid=major,
    grid style={black!12, line width=0.3pt},
    every axis plot/.append style={line width=0.9pt, line join=round},
    legend style={font=\scriptsize, draw=none, fill=white, fill opacity=0.85,
                  text opacity=1, legend columns=3, column sep=6pt,
                  at={(0.98,0.04)}, anchor=south east},
    legend cell align=left,
    legend image code/.code={\draw[#1, line width=1.2pt] (0cm,0cm) -- (0.4cm,0cm);},
]

\addplot[draw=none, fill=colorG3PO, fill opacity=0.15, forget plot] coordinates {
    (0, 0.5614)
    (5, 0.5901)
    (10, 0.6124)
    (15, 0.6277)
    (20, 0.6500)
    (25, 0.6906)
    (30, 0.7200)
    (35, 0.7338)
    (40, 0.7382)
    (45, 0.7595)
    (50, 0.7757)
    (55, 0.7594)
    (60, 0.7656)
    (65, 0.7817)
    (70, 0.7772)
    (75, 0.7808)
    (80, 0.7883)
    (85, 0.7790)
    (90, 0.7630)
    (95, 0.7493)
    (100, 0.7517)
    (105, 0.7527)
    (110, 0.7587)
    (115, 0.7630)
    (120, 0.7503)
    (125, 0.7417)
    (130, 0.7554)
    (135, 0.7614)
    (140, 0.7559)
    (145, 0.7704)
    (150, 0.7984)
    (155, 0.7865)
    (160, 0.7740)
    (165, 0.7821)
    (170, 0.7689)
    (175, 0.7569)
    (180, 0.7697)
    (185, 0.7863)
    (190, 0.7963)
    (195, 0.7932)
    (200, 0.7855)
    (205, 0.8025)
    (210, 0.7994)
    (215, 0.8011)
    (220, 0.8177)
    (225, 0.8160)
    (230, 0.7841)
    (230, 0.7559)
    (225, 0.7890)
    (220, 0.7973)
    (215, 0.7789)
    (210, 0.7756)
    (205, 0.7725)
    (200, 0.7645)
    (195, 0.7668)
    (190, 0.7687)
    (185, 0.7587)
    (180, 0.7453)
    (175, 0.7281)
    (170, 0.7461)
    (165, 0.7529)
    (160, 0.7460)
    (155, 0.7635)
    (150, 0.7666)
    (145, 0.7446)
    (140, 0.7341)
    (135, 0.7386)
    (130, 0.7246)
    (125, 0.7183)
    (120, 0.7297)
    (115, 0.7370)
    (110, 0.7313)
    (105, 0.7323)
    (100, 0.7283)
    (95, 0.7257)
    (90, 0.7370)
    (85, 0.7560)
    (80, 0.7567)
    (75, 0.7492)
    (70, 0.7528)
    (65, 0.7533)
    (60, 0.7394)
    (55, 0.7356)
    (50, 0.7443)
    (45, 0.7355)
    (40, 0.7168)
    (35, 0.7112)
    (30, 0.6950)
    (25, 0.6644)
    (20, 0.6300)
    (15, 0.5973)
    (10, 0.5826)
    (5, 0.5649)
    (0, 0.5386)
};
\addplot[color=colorG3PO, mark=none] coordinates {
    (0, 0.5500)
    (5, 0.5775)
    (10, 0.5975)
    (15, 0.6125)
    (20, 0.6400)
    (25, 0.6775)
    (30, 0.7075)
    (35, 0.7225)
    (40, 0.7275)
    (45, 0.7475)
    (50, 0.7600)
    (55, 0.7475)
    (60, 0.7525)
    (65, 0.7675)
    (70, 0.7650)
    (75, 0.7650)
    (80, 0.7725)
    (85, 0.7675)
    (90, 0.7500)
    (95, 0.7375)
    (100, 0.7400)
    (105, 0.7425)
    (110, 0.7450)
    (115, 0.7500)
    (120, 0.7400)
    (125, 0.7300)
    (130, 0.7400)
    (135, 0.7500)
    (140, 0.7450)
    (145, 0.7575)
    (150, 0.7825)
    (155, 0.7750)
    (160, 0.7600)
    (165, 0.7675)
    (170, 0.7575)
    (175, 0.7425)
    (180, 0.7575)
    (185, 0.7725)
    (190, 0.7825)
    (195, 0.7800)
    (200, 0.7750)
    (205, 0.7875)
    (210, 0.7875)
    (215, 0.7900)
    (220, 0.8075)
    (225, 0.8025)
    (230, 0.7700)
};
\addlegendentry{G3PO}

\addplot[draw=none, fill=colorGRPO, fill opacity=0.15, forget plot] coordinates {
    (0, 0.5728)
    (5, 0.5866)
    (10, 0.6348)
    (15, 0.6947)
    (20, 0.6238)
    (25, 0.6967)
    (30, 0.7807)
    (35, 0.7980)
    (40, 0.6923)
    (45, 0.7829)
    (50, 0.7620)
    (55, 0.7322)
    (60, 0.7828)
    (65, 0.7004)
    (70, 0.6989)
    (75, 0.7551)
    (80, 0.7483)
    (85, 0.7974)
    (90, 0.7152)
    (95, 0.6688)
    (100, 0.8284)
    (105, 0.7624)
    (110, 0.7489)
    (115, 0.7144)
    (120, 0.7142)
    (125, 0.7531)
    (130, 0.7006)
    (135, 0.8091)
    (140, 0.7225)
    (145, 0.7529)
    (150, 0.7750)
    (155, 0.8674)
    (160, 0.7619)
    (165, 0.7476)
    (170, 0.8072)
    (175, 0.6576)
    (180, 0.8060)
    (185, 0.6719)
    (190, 0.7343)
    (195, 0.8126)
    (200, 0.8005)
    (205, 0.8340)
    (210, 0.7804)
    (215, 0.7843)
    (220, 0.7594)
    (225, 0.7857)
    (230, 0.7679)
    (230, 0.6906)
    (225, 0.7295)
    (220, 0.7075)
    (215, 0.7086)
    (210, 0.7292)
    (205, 0.7614)
    (200, 0.7259)
    (195, 0.7451)
    (190, 0.6862)
    (185, 0.6317)
    (180, 0.7328)
    (175, 0.6125)
    (170, 0.7668)
    (165, 0.6772)
    (160, 0.6769)
    (155, 0.8193)
    (150, 0.7308)
    (145, 0.6808)
    (140, 0.6435)
    (135, 0.7247)
    (130, 0.6158)
    (125, 0.6768)
    (120, 0.6568)
    (115, 0.6574)
    (110, 0.6772)
    (105, 0.6773)
    (100, 0.7435)
    (95, 0.6241)
    (90, 0.6631)
    (85, 0.7309)
    (80, 0.6805)
    (75, 0.6743)
    (70, 0.6259)
    (65, 0.6475)
    (60, 0.6990)
    (55, 0.6459)
    (50, 0.7163)
    (45, 0.7259)
    (40, 0.6454)
    (35, 0.7111)
    (30, 0.7214)
    (25, 0.6222)
    (20, 0.5751)
    (15, 0.6224)
    (10, 0.5835)
    (5, 0.5210)
    (0, 0.5319)
};
\addplot[color=colorGRPO, mark=none] coordinates {
    (0, 0.5524)
    (5, 0.5538)
    (10, 0.6091)
    (15, 0.6585)
    (20, 0.5995)
    (25, 0.6595)
    (30, 0.7511)
    (35, 0.7545)
    (40, 0.6689)
    (45, 0.7544)
    (50, 0.7391)
    (55, 0.6890)
    (60, 0.7409)
    (65, 0.6739)
    (70, 0.6624)
    (75, 0.7147)
    (80, 0.7144)
    (85, 0.7641)
    (90, 0.6891)
    (95, 0.6464)
    (100, 0.7860)
    (105, 0.7198)
    (110, 0.7130)
    (115, 0.6859)
    (120, 0.6855)
    (125, 0.7150)
    (130, 0.6582)
    (135, 0.7669)
    (140, 0.6830)
    (145, 0.7169)
    (150, 0.7529)
    (155, 0.8434)
    (160, 0.7194)
    (165, 0.7124)
    (170, 0.7870)
    (175, 0.6351)
    (180, 0.7694)
    (185, 0.6518)
    (190, 0.7102)
    (195, 0.7789)
    (200, 0.7632)
    (205, 0.7977)
    (210, 0.7548)
    (215, 0.7465)
    (220, 0.7335)
    (225, 0.7576)
    (230, 0.7293)
};
\addlegendentry{GRPO}

\addplot[draw=none, fill=colorDAPO, fill opacity=0.15, forget plot] coordinates {
    (0, 0.4795)
    (5, 0.5030)
    (10, 0.4874)
    (15, 0.5268)
    (20, 0.5262)
    (25, 0.5876)
    (30, 0.6147)
    (35, 0.6606)
    (40, 0.6444)
    (45, 0.6457)
    (50, 0.6941)
    (55, 0.6482)
    (60, 0.6813)
    (65, 0.6834)
    (70, 0.6693)
    (75, 0.6537)
    (80, 0.6662)
    (85, 0.6974)
    (90, 0.6762)
    (95, 0.6294)
    (100, 0.6686)
    (105, 0.6591)
    (110, 0.6401)
    (115, 0.6955)
    (120, 0.6723)
    (125, 0.6372)
    (130, 0.6538)
    (135, 0.6502)
    (140, 0.6513)
    (145, 0.6716)
    (150, 0.7158)
    (155, 0.6993)
    (160, 0.6164)
    (165, 0.6848)
    (170, 0.6932)
    (175, 0.6388)
    (180, 0.6758)
    (185, 0.6596)
    (190, 0.7021)
    (195, 0.7183)
    (200, 0.6669)
    (205, 0.7041)
    (210, 0.7006)
    (215, 0.7022)
    (220, 0.7375)
    (225, 0.7104)
    (230, 0.6835)
    (230, 0.6267)
    (225, 0.6686)
    (220, 0.7116)
    (215, 0.6407)
    (210, 0.6548)
    (205, 0.6800)
    (200, 0.6230)
    (195, 0.6691)
    (190, 0.6428)
    (185, 0.6162)
    (180, 0.6487)
    (175, 0.6055)
    (170, 0.6375)
    (165, 0.6486)
    (160, 0.5840)
    (155, 0.6435)
    (150, 0.6553)
    (145, 0.6110)
    (140, 0.6133)
    (135, 0.5912)
    (130, 0.6331)
    (125, 0.6017)
    (120, 0.6139)
    (115, 0.6354)
    (110, 0.5806)
    (105, 0.6317)
    (100, 0.6215)
    (95, 0.6084)
    (90, 0.6444)
    (85, 0.6470)
    (80, 0.6380)
    (75, 0.6131)
    (70, 0.6251)
    (65, 0.6423)
    (60, 0.6279)
    (55, 0.6017)
    (50, 0.6366)
    (45, 0.6092)
    (40, 0.5835)
    (35, 0.6303)
    (30, 0.5836)
    (25, 0.5521)
    (20, 0.5023)
    (15, 0.4829)
    (10, 0.4397)
    (5, 0.4473)
    (0, 0.4322)
};
\addplot[color=colorDAPO, mark=none] coordinates {
    (0, 0.4559)
    (5, 0.4752)
    (10, 0.4636)
    (15, 0.5049)
    (20, 0.5142)
    (25, 0.5698)
    (30, 0.5992)
    (35, 0.6455)
    (40, 0.6140)
    (45, 0.6274)
    (50, 0.6654)
    (55, 0.6250)
    (60, 0.6546)
    (65, 0.6628)
    (70, 0.6472)
    (75, 0.6334)
    (80, 0.6521)
    (85, 0.6722)
    (90, 0.6603)
    (95, 0.6189)
    (100, 0.6451)
    (105, 0.6454)
    (110, 0.6103)
    (115, 0.6654)
    (120, 0.6431)
    (125, 0.6195)
    (130, 0.6435)
    (135, 0.6207)
    (140, 0.6323)
    (145, 0.6413)
    (150, 0.6855)
    (155, 0.6714)
    (160, 0.6002)
    (165, 0.6667)
    (170, 0.6654)
    (175, 0.6222)
    (180, 0.6622)
    (185, 0.6379)
    (190, 0.6725)
    (195, 0.6937)
    (200, 0.6449)
    (205, 0.6921)
    (210, 0.6777)
    (215, 0.6715)
    (220, 0.7245)
    (225, 0.6895)
    (230, 0.6551)
};
\addlegendentry{DAPO}

\end{axis}
\end{tikzpicture}
    \caption{Outcome training reward of G3PO, GRPO and DAPO on MATH (shaded: confidence interval). G3PO converges more smoothly and with lower variance than GRPO, and reaches a higher final reward than DAPO.}
    \label{fig:train_dyn}
\end{figure}
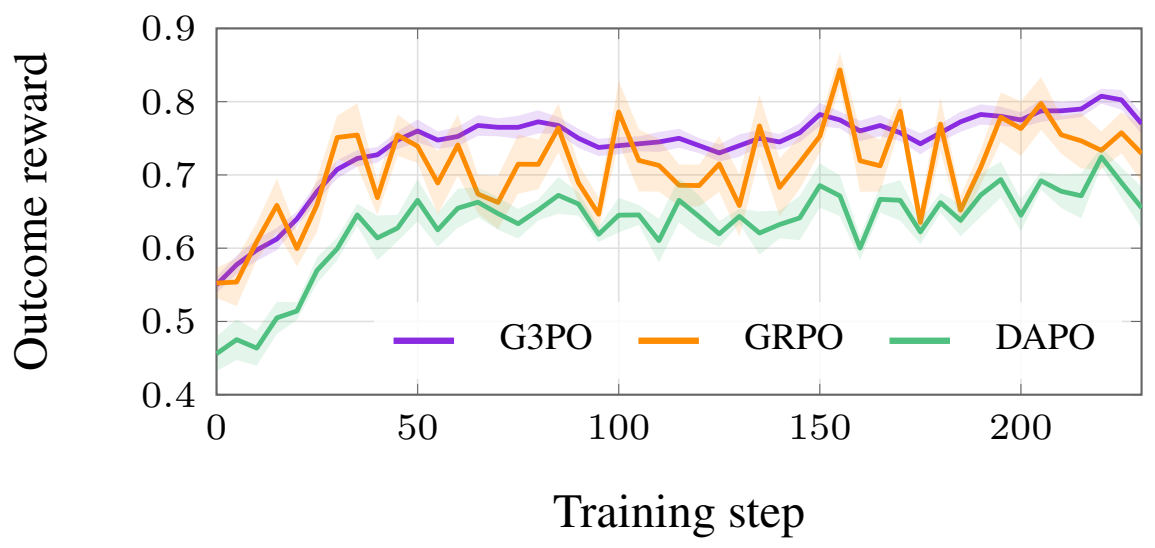

Figure~\ref{fig:train_dyn} shows the outcome reward across optimisation steps during training. G3PO demonstrates a distinct advantage in both sample efficiency and training stability. While GRPO achieves high peak rewards intermittently, it suffers from severe training instability, characterised by heavy oscillations and high variance (indicated by the wide shaded confidence intervals). This volatility is a known challenge in policy gradient methods when optimising for sparse or verifiable rewards.

\section{\texorpdfstring{\add{Rubric-Based Evaluation}}{Rubric-Based Evaluation}}\label{app:rar}
\add{The benchmarks of Section~\ref{sec:results} all have short answers, such as a multiple-choice option or a number, that can be checked against a reference. To assess whether the gains of G3PO carry over to open-ended answers, we evaluate the medical-reasoning models on all 2,242 prompts of the RaR-Medicine test split~\cite{gunjal2026rubrics}. Each prompt comes with an expert reference answer and an instance-specific rubric, a list of criteria that a good answer should meet, each tagged as Essential, Important, Optional or Pitfall and weighted from 1 to 5 (pitfalls carry negative weights in the released data).}

\paragraph{\add{Generation.}} \add{The models trained for medical reasoning (Table~\ref{tab:medical_benchmarks}) answer each prompt zero-shot, without any training on RaR-Medicine, with greedy decoding and at most 1,024 new tokens. We strip the reasoning trace and pass only the content of the \texttt{<answer>} tags to the judge, since the rubrics describe the answer and not the reasoning; a generation truncated before its answer is judged as empty.}

\paragraph{\add{Judging.}} \add{Following the HealthBench protocol~\cite{arora2025healthbench}, which RaR uses for this data, Claude Opus~5~\cite{anthropic2026opus5} grades one criterion per call at temperature 0. The judge sees the question, the reference answer, the candidate answer and the criterion with its point value, and returns whether the criterion is met (prompt in Figure~\ref{fig:rar_judge}). A call that does not return a valid verdict is retried up to five times.}

\paragraph{\add{Scoring.}} \add{The score of an answer is the weight of the criteria it meets divided by the total positive weight of its rubric, and the reported score is the mean over prompts, clipped to $[0,1]$. The released pitfall criteria are mostly phrased as an error to avoid (\emph{``Does not mention X''}), while RaR describes them as positively phrased and weights them at $0.9$ times an Essential criterion. We therefore grade each pitfall as written, with a positive weight of $0.9\times5=4.5$, so that rewriting the criterion text cannot flip its meaning. One test criterion with a missing weight takes the modal weight of its tier.}

\paragraph{\add{Results.}} \add{G3PO obtains an average rubric score of 48.1\%, against 45.4\% for GRPO and 19.1\% for GOPO (Table~\ref{tab:rar}). It thus retains its advantage over GRPO observed on the multiple-choice benchmarks of Table~\ref{tab:medical_benchmarks}, suggesting that its gains are not tied to answers that can be verified exactly.}

\begin{table}[htbp]
\centering
\small
\begin{tabular}{lc}
\toprule
& \add{\textbf{Avg.\ rubric score}\,$\uparrow$} \\
\midrule
\add{GRPO} & \add{45.4} \\
\add{GOPO} & \add{19.1} \\
\rowcolor{g3poTint}
\add{\textbf{G3PO (ours)}} & \add{\best{48.1}} \\
\bottomrule
\end{tabular}
\caption{\add{Average rubric score (\%) on the RaR-Medicine test split, zero-shot, judged by Claude Opus~5.}}
\label{tab:rar}
\end{table}

\begin{promptbox}[breakable, colupper=rev]{Judge prompt for RaR-Medicine (one call per criterion)}
Your job is to look at a medical question, a reference answer, and a candidate response, and score the candidate response on how well it satisfies a single rubric item.

\smallskip
\# Question\\
\{question\}

\smallskip
\# Reference answer (written by an expert; the candidate response does not have to match it word for word)\\
\{reference\_answer\}

\smallskip
\# Candidate response\\
\{response\}

\smallskip
\# Rubric item\\
{[}\{points\} points{]} \{criterion\}

\smallskip
\# Instructions\\
Return a json object with the following fields: "explanation" and "criteria\_met".\\
- The "explanation" field should be a short string explaining why the candidate response does or does not meet the rubric item.\\
- The "criteria\_met" field should be a boolean indicating whether the candidate response meets the rubric item. If the rubric item has multiple sentences or criteria, consider all of them: if any one of them is not met, return false. Only return true if all of them are met.\\
- One important exception is that if the rubric item says "such as", "for example", or "including", the response does not have to include every example listed in order to meet the criterion. If the item says "States that oral iron supplements can lead to unpleasant gastrointestinal side effects such as nausea, vomiting, and constipation", and the response just says oral iron can cause unpleasant gastrointestinal side effects such as cramps, the criterion is still met.\\
- Judge the rubric item literally, as written: report whether the item is met, not whether the response is good overall. Some items describe something undesirable, or are phrased as an omission ("Does not mention X"). For an item phrased as "Does not mention X", it is met when the response genuinely does not mention X. For an item like "Recommends a medication other than ketoconazole", it is NOT met when the response correctly recommends ketoconazole.\\
- Judge only the content of the candidate response. Do not reward it for restating the question, and do not penalise stylistic differences from the reference answer.\\
- Return only the json object, with no other text.

\smallskip
\# Example\\
\textasciigrave\textasciigrave\textasciigrave json\\
\{\\
\phantom{xx}"explanation": "The response identifies DiGeorge syndrome and attributes the hypocalcemia to parathyroid hypoplasia, which is what the criterion asks for.",\\
\phantom{xx}"criteria\_met": true\\
\}\\
\textasciigrave\textasciigrave\textasciigrave
\end{promptbox}
\par\nobreak\noindent\begin{minipage}{\columnwidth}
\captionof{figure}{\add{Prompt given to Claude Opus~5 to grade one rubric criterion of a RaR-Medicine answer.}}
\label{fig:rar_judge}
\end{minipage}

\begin{figure}[!t]
\centering
\large
\begin{minipage}{\columnwidth}

\begin{promptbox}{Prompt used for RoB Assessment}
\textbf{system\_prompt:} "You're an expert in estimating risk of bias in clinical trial text. Kindly address the given query. Answer with the best option directly. You FIRST think about the reasoning process as an internal monologue and then provide the user with the answer. The reasoning process MUST BE enclosed within <think> and </think> tags."

\smallskip
\textbf{user\_prompt\_template:} "Input:\\
Article: \{article\}\\[3pt]
Question: Based on the given trial article, what is the risk of bias for the following Comparison and Outcome?\\[3pt]
Comparison: \{comparator\}\\
Outcome: \{outcome\}\\[3pt]
The bias you have to assess is defined as follows:\\
\{bias\}\\[3pt]
Think about it step by step and output your reasoning between <think></think> tags. Explain thoroughly how you performed the assessment. The final response should be in the following format:\\[3pt]
<think>\\
...\\
</think>\\
<answer>\\
risk: high|low|moderate\\
</answer>"
\end{promptbox}

\begin{promptbox}{Prompt used for Medical Reasoning}
\textbf{system\_prompt:} "You are a medical expert. Please reason step by step, put your reasoning in the <think><think> tags and put the letter corresponding to your final answer within \textbackslash boxed\{\} in the <answer></answer> tags."

\smallskip
\textbf{user\_prompt\_template:} "Question: \{question\}\\
\{options\}"
\end{promptbox}

\begin{promptbox}{Prompt used for Mathematics}
\textbf{system\_prompt:} "Please reason step by step, put your reasoning in the <think><think> tags and put your final answer within \textbackslash boxed\{\} in the <answer></answer> tags."

\smallskip
\textbf{user\_prompt\_template:} "\{problem\}"
\end{promptbox}

\end{minipage}
\caption{Prompts used across the various domains to guide policy behaviour during training and for inference.}
\label{fig:prompt_examples}
\end{figure}

\section{Additional Background on Risk of Bias Assessment in Medical Systematic Reviews}

A Risk of Bias (RoB) assessment is a critical component of systematic reviews, which are structured syntheses of medical literature designed to collect, appraise, and summarise evidence on a specific clinical question. In this setting, RoB evaluation is used to assess the internal validity of included studies and to identify factors that can distort reported effects, such as funding bias arising from industry sponsorship or outcome reporting bias, where favourable results are selectively emphasised while negative findings are under-reported. Medical systematic reviews commonly rely on Cochrane-developed tools for randomised and non-randomised trials \cite{sterne2019, robins2025}. Cochrane plays a central role in this process by developing widely used methodological standards and bias-assessment frameworks \cite{higgins2024}, and its RoB tools provide a domain-based set of questions that guide reviewers in judging the likelihood of bias in each study. These domain-level judgments are then combined to form an overall assessment of the study’s overall risk of bias. The existence of these step-level results motivated the use of verifiable Process Reward Models (VPRMs) \cite{pronesti2026beyond}. 

\section{Prompts}
Figure~\ref{fig:prompt_examples} depicts the prompts used for training and inference. While varying slightly between domain applications, we emphasise the continued simplicity between them, as applying more specific prompts would often constrain the answer, leading to formatting or other scoring errors.

\section{Training Examples}\label{algApp}
Figure~\ref{visual_demonstration} illustrates several training generations, their correctness scores, the corresponding PRM evaluations, and the advantage assignments produced by G3PO. Notably, no completion with correctness $=0.0$ receives a higher advantage than any completion with correctness $=1.0$, regardless of its PRM score. This preserves the correctness ordering imposed by verifiability while still allowing the model to differentiate among trajectories using the preference signal.

\section{Scientific Artefacts and Licensing}\label{app:licensing}

In this work, we used the following scientific artefacts. LLaMa 3.1 is licensed under a commercial license\footnote{\url{https://llama.meta.com/doc/overview}}. Qwen2.5, Qwen2.5-Math and their derivatives are licensed under the Apache 2.0 license\footnote{\url{https://qwenlm.github.io/blog/qwen3}}.  DeepSeek models are licensed under the MIT license\footnote{\url{https://api-docs.deepseek.com/news/news250120}}. NFT-7B is licensed under the Nvidia non-commercial license\footnote{\url{https://huggingface.co/nvidia/NFT-7B/blob/main/LICENSE}}. Mining text and data from the Cochrane library is permitted for non-commercial research through the Wiley API\footnote{\url{https://www.cochranelibrary.com/help/access}}. \add{RaR-Medicine~\cite{gunjal2026rubrics} is publicly released on Hugging Face.} \add{MATH~\cite{hendrycks2021measuring_mathdataset}, GSM8K~\cite{cobbe_gsm8k}, OlympiadBench~\cite{olympiadbench} and PRM800K~\cite{Lightman_math500}, which defines the MATH500 split, are released under the MIT license, and the AIME 2024--2026 sets~\cite{aime24, aime25, aime26} under Apache 2.0.} The usage of the listed artefacts is consistent with their licenses.

\begin{figure*}[t]
    \centering
    \input{diagrams/visual_example}
    \caption{{\bfseries Qualitative Comparison of G3PO Outputs,} \add{four of the $G=8$ completions for one MATH prompt at the final training step, arranged by verifier outcome (columns) and PRM score (rows); excerpts, with \LaTeX{} rendered and code comments elided ([\dots]),} with green highlights marking the reasoning fragments that receive positive credit and red highlights marking fragments that are penalised by the PRM. Green spans typically correspond to structurally sound steps such as coefficient matching, the quadratic reformulation, and discriminant-based solvability checks. Red spans mark weaker or overly computational steps, including less justified search-based reasoning or logic that does not meaningfully improve the derivation. \add{The strip at the bottom gives the G3PO rank and advantage of every completion in the group: the correct completion (d) outranks the incorrect (b) despite its much lower PRM score.}}
    \label{visual_demonstration}
\end{figure*}

\end{document}